%% file: neurips_2026.tex
\documentclass{article}

    \PassOptionsToPackage{numbers, sort}{natbib}
\input{math_commands.tex}
 \usepackage[main, final]{neurips_2026}
\usepackage[utf8]{inputenc} % allow utf-8 input
\usepackage[T1]{fontenc}    % use 8-bit T1 fonts
\usepackage[dvipsnames,svgnames,table]{xcolor}
\usepackage{pifont}
\usepackage{fontawesome5}

\definecolor{crimson}{HTML}{DC143C}

\usepackage{amsmath, amssymb, amsfonts, amsthm, mathtools}
\usepackage{bm}           % For bold math
\usepackage{nicefrac}     % For compact fractions
\usepackage{dsfont}       % For \mathds (used by \indicator)
\usepackage{bbm}          % \mathbbm (kept from NeurIPS template)
\usepackage{xparse}

\usepackage{graphicx}
\usepackage{booktabs}       % professional-quality tables
\usepackage{longtable}
\usepackage{multirow}
\usepackage{makecell}
\usepackage{array}
\usepackage{caption}
\usepackage{subcaption}
\usepackage{wrapfig}
\usepackage{float}
\usepackage{scalerel}
\usepackage{colortbl}
\usepackage{tablefootnote}
\usepackage{adjustbox}
\usepackage{dblfloatfix}    % fixes two-column floats
\usepackage[algo2e, linesnumbered, ruled, vlined]{algorithm2e}
\usepackage{algorithm}
\usepackage{algorithmic}

\makeatletter
\@namedef{ver@everyshi.sty}{}
\makeatother

\usepackage{microtype}      % microtypography
\usepackage{xurl}           % better URL line breaking (replaces url)
\usepackage{enumitem}
\usepackage[normalem]{ulem} % Provides \sout
\usepackage{soul}           % Provides \st, \ul
\usepackage[textsize=tiny]{todonotes}
\usepackage{tocloft}        % table of contents spacing
\usepackage[pagebackref=true,linktoc=all]{hyperref}
\usepackage[capitalize,noabbrev]{cleveref} % Must load after hyperref

\usepackage{kotex}

\definecolor{darkblue}{rgb}{0.0,0.0,0.65}
\definecolor{darkred}{rgb}{0.65,0.0,0.0}
\definecolor{darkgreen}{rgb}{0.0,0.5,0.0}
\definecolor{pinegreen}{RGB}{1, 121, 111}
\definecolor{tab:blue}{RGB}{31,119,180}
\definecolor{tab:red}{RGB}{214,39,40}
\definecolor{tab:green}{RGB}{44,160,44}
\definecolor{tab:orange}{RGB}{255,127,14}
\hypersetup{
    colorlinks=true,
    citecolor=darkblue,
    linkcolor=darkred,
    filecolor=darkblue,
    urlcolor=darkblue,
}

\newtheoremstyle{bastion-remark}
  {0.8em}      % space above
  {0.6em}      % space below
  {\itshape}   % body font (italic)
  {0pt}        % indent
  {\bfseries}  % header font (bold)
  {.}          % punctuation after header
  { }          % space after header (single space)
  {\thmname{#1}~\thmnumber{#2}\thmnote{ (\normalfont\itshape #3)}}
  
\theoremstyle{plain}
\newtheorem{theorem}{Theorem}[section]
\newtheorem{proposition}[theorem]{Proposition}
\newtheorem{lemma}[theorem]{Lemma}

\theoremstyle{definition}

\theoremstyle{bastion-remark}

\newcommand{\xmark}{\textcolor{DarkRed}{\ding{55}}}

\newcommand{\algo}{\textsc{Bastion}\xspace}

\NewDocumentCommand{\bastionicon}{}{%
  \smash{\raisebox{-0.38cm}{%
    \includegraphics[height=1.35cm]{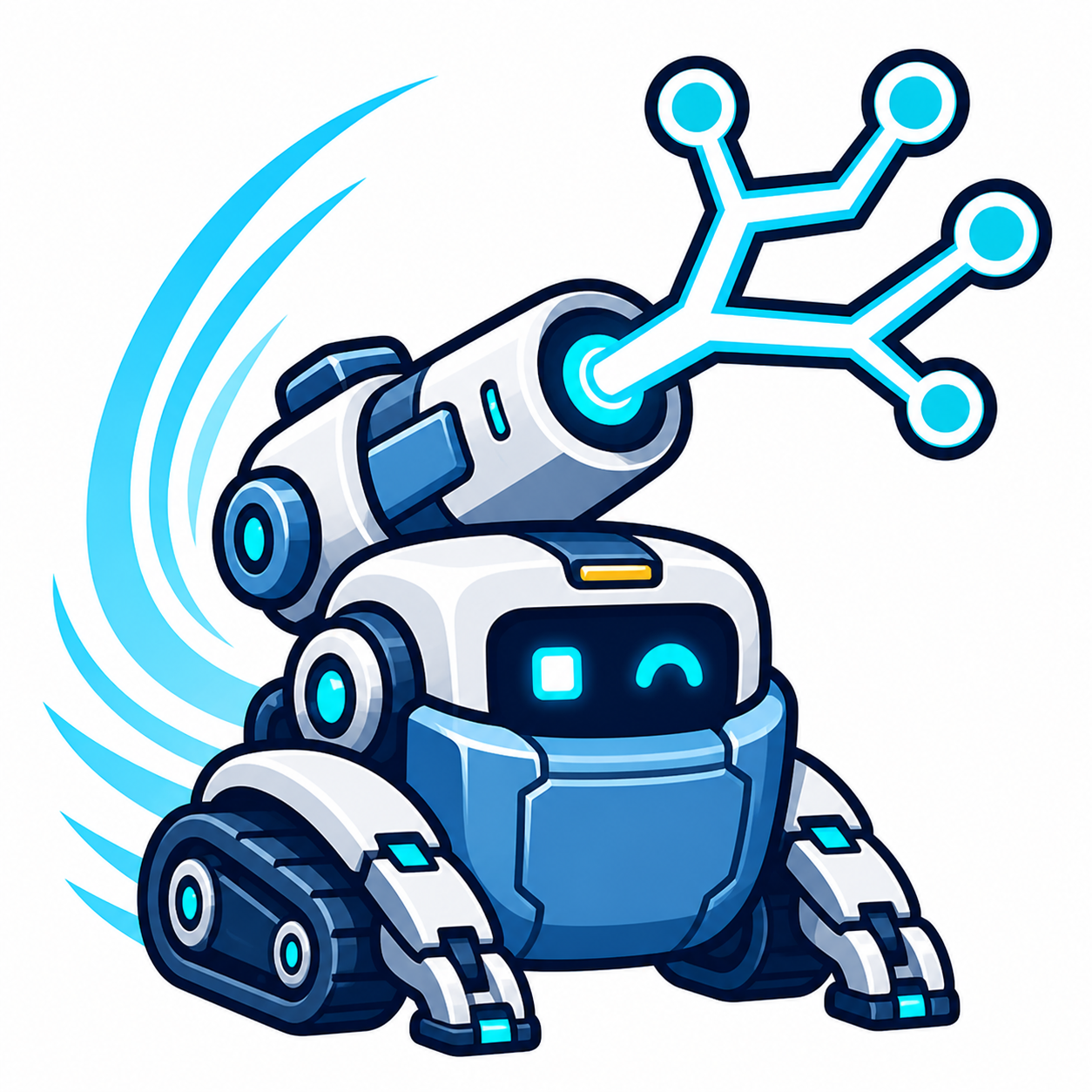}%
  }}%
}

\title{
\bastionicon\hspace{0.15cm}BASTION: Budget-Aware Speculative Decoding\\with Tree-structured Block Diffusion Drafting}
\author{
  Soowon Oh\textsuperscript{\scalebox{0.7}{$\spadesuit$}$\diamond$}\thanks{Both authors contribute equally to this work.}, \;
  Nam Cao\textsuperscript{\scalebox{0.7}{$\spadesuit$}}\footnotemark[1], \;
  Yujin Kim\textsuperscript{\scalebox{0.7}{$\spadesuit$}}, \;
  Hojung Jung\textsuperscript{\scalebox{0.7}{$\spadesuit$}}, \;
  Huzama Ahmad\textsuperscript{\scalebox{0.7}{$\spadesuit$}},\\
  \textbf{Sangmin Bae}\textsuperscript{\scalebox{0.7}{$\spadesuit$}}\thanks{Corresponding authors},\;
  \textbf{Se-Young Yun}\textsuperscript{\scalebox{0.7}{$\spadesuit$}}\footnotemark[2]\\
  \textsuperscript{\scalebox{0.7}{$\spadesuit$}}KAIST AI \quad \textsuperscript{$\diamond$}Samsung Advanced Institute of Technology\\
  \footnotesize{\{osw5144,hainam,yunseyoung\}@kaist.ac.kr}
}

\begin{document}

\maketitle

\input{tex/00_abstract}

\input{tex/01_introduction}

\input{tex/03_prelim}

\input{tex/04_method_new}
\input{tex/05_experiments}

\input{tex/06_analysis}
\input{tex/06_conclusion}

%%%%%%%%%%%%%%%%%%%%%%%%%%%%%%%%%%%%%%%%%%%%%%%%%%%%%%%%%%%%

\clearpage

\section*{Acknowledgements}
This work was partly supported by the Institute for Information \& Communications Technology Planning \& Evaluation (IITP) grant funded by the Korean government (MSIT) ([RS-2019-II190075, Artificial Intelligence Graduate School Program (KAIST), 5\%], [RS-2024-00457882, National AI Research Lab Project, 5\%].

\bibliographystyle{plainnat}
\bibliography{reference}

\input{tex/07_appendix}

% \clearpage
% \input{checklist}

\end{document}

%% file: math_commands.tex
\usepackage{amsmath,amsfonts,bm}

\def\eqref#1{equation~\ref{#1}}
\def\Eqref#1{Equation~\ref{#1}}
\def\1{\bm{1}}

\DeclareMathAlphabet{\mathsfit}{\encodingdefault}{\sfdefault}{m}{sl}
\SetMathAlphabet{\mathsfit}{bold}{\encodingdefault}{\sfdefault}{bx}{n}

%% file: tex/00_abstract.tex
\begin{abstract}
\label{abstract}
% Block-diffusion drafters have recently shown strong potential for speculative decoding by predicting multiple future-token distributions in a single forward pass. To exploit these parallel predictions, tree-based verification can check multiple plausible continuations instead of committing to a single greedy draft path, often increasing the number of accepted tokens per decoding cycle. However, existing tree-based approaches typically rely on fixed tree topologies or static verification budgets, even though the best budget depends on the local draft distribution, context length, target model, and hardware runtime. We propose a system-aware adaptive tree framework for block-diffusion speculative decoding. At each decoding cycle, our method builds a tree toward high-probability draft paths and selects the verification budget that maximizes an online speedup estimate. This estimate combines a drafter-based acceptance surrogate with an analytical verifier-latency model calibrated from observed runtimes. The framework requires no additional training of the drafter or target model and preserves the target model's decoding rule. 
% Across target models, benchmarks, decoding temperatures, and GPU platforms, our method achieves up to a $6.93\times$ wall-clock speedup over standard autoregressive decoding. This corresponds to approximately $40\%$ higher speedup than Dflash, the strongest existing block-diffusion drafter, and is obtained without per-setting budget tuning.

Block-diffusion drafters have recently emerged as a powerful alternative for speculative decoding by predicting multiple future-token distributions in a single parallel step. However, since these parallel predictions are sampled from position-wise marginals rather than fully conditioned sequences, committing to a single greedy path often fails to capture the target model's preferred trajectory. To address this, we propose \textbf{\algo}, a \textbf{b}udget-\textbf{a}ware \textbf{s}peculative decoding framework with \textbf{t}ree-based diffus\textbf{ion} drafting. Unlike existing methods that rely on static tree topologies, \algo dynamically constructs query-dependent trees by balancing draft quality against hardware constraints. Our framework integrates three synergistic components: (1) an \textit{acceptance surrogate} that estimates expected accepted length via path confidence, (2) an \textit{online latency estimator} that calibrates a hardware-aware roofline model, and (3) an \textit{adaptive best-first expansion} that grows the tree until marginal gains no longer justify incremental verification costs. \algo is training-free, preserves the target model's distribution, and requires no per-setting tuning. Across diverse benchmarks and GPU architectures, \algo achieves up to a \textbf{$6.61\times$} speedup over standard autoregressive decoding, outperforming state-of-the-art block-diffusion baselines by \textbf{$39\%$}.
\begin{center}
    \faGithub \;Code: \href{https://github.com/kaist-ai-osi-lab/BASTION}{https://github.com/kaist-ai-osi-lab/BASTION}
\end{center}
\vspace{-2.5em}
\end{abstract}

%% file: tex/01_introduction.tex
\input{figures/overview}

% \clearpage
\vspace{-5pt}
\section{Introduction}
\label{sec:introduction}

% \vspace{-1pt}
% Recent large language models~(LLMs) demonstrate impressive reasoning capabilities~\citep{grattafiori2024llama, comanici2025gemini, yang2025qwen3, agarwal2025gpt, deepseekai2026deepseekv4}, but their serving costs remain high because autoregressive~(AR) decoding requires a separate target-model forward pass for every generated token. \textit{Speculative decoding} mitigates this sequential bottleneck by using a fast draft generator---either a smaller auxiliary model~\citep{chen2023accelerating, leviathan2023fast, kim2023speculative, miao2024specinfer} or a self-drafting mechanism~\citep{bae2023fast, elhoushi2024layerskip, cai2024medusa, li2024eagle, zhang2024draft}---to propose multiple candidate tokens. These are verified in parallel by the target model using an exact acceptance rule~\citep{chen2023accelerating, leviathan2023fast} (or a relaxed rule~\citep{cai2024medusa, jang2024lantern, bachmann2025judge}). This preserves the target model's output distribution while shifting generation from token-by-token decoding to batched verification. In practice, the resulting speedup is determined by the acceptance rate versus the auxiliary cost; thus, effective speculative decoding requires carefully balancing draft quality against generation overhead.
% \looseness=-1

Recent large language models~(LLMs) demonstrate impressive reasoning capabilities~\citep{grattafiori2024llama, comanici2025gemini, yang2025qwen3, agarwal2025gpt, deepseekai2026deepseekv4}, but their serving costs remain high because autoregressive~(AR) decoding requires a separate target-model forward pass for every generated token. \textit{Speculative decoding} mitigates this bottleneck by using a fast drafter---either a smaller auxiliary model~\citep{chen2023accelerating, leviathan2023fast, kim2023speculative, miao2024specinfer} or a self-drafting mechanism~\citep{bae2023fast, elhoushi2024layerskip, cai2024medusa, li2024eagle, zhang2024draft}---to propose multiple candidate tokens, which the target model verifies in parallel with an exact acceptance rule~\citep{chen2023accelerating, leviathan2023fast} or a relaxed variant~\citep{cai2024medusa, jang2024lantern, bachmann2025judge}. Under exact acceptance, this preserves the target distribution while shifting generation from token-by-token decoding to batched verification, with speedup governed by the acceptance rate relative to drafting and verification overhead.
\looseness=-1

% Most prior speculative decoding methods optimize this efficiency trade-off within an autoregressive drafting paradigm~\citep{leviathan2023fast, sadhukhan2024magicdec, elhoushi2024layerskip, li2025eagle}, designing lightweight yet context-aware drafters to preserve high predictive accuracy. State-of-the-art methods such as EAGLE-3~\citep{li2025eagle} deliver practical speedups in the 2--3$\times$ range. Nevertheless, because their drafting phase remains inherently sequential, generating a long candidate block requires repeated draft-model steps, each conditioned on earlier tokens. As a result, even a cheap AR drafter underutilizes available parallel compute, making the draft stage costly as the proposed block length grows.
% \looseness=-1

Most prior methods optimize this trade-off within an autoregressive drafting paradigm~\citep{leviathan2023fast, sadhukhan2024magicdec, elhoushi2024layerskip, li2025eagle}, designing lightweight yet context-aware drafters. State-of-the-art methods such as EAGLE-3~\citep{li2025eagle} deliver practical speedups in the 2--3$\times$ range. However, because their drafting phase remains sequential, longer candidate blocks require repeated draft-model steps conditioned on previous draft tokens, making the draft stage increasingly costly and underutilizing available parallel compute.
\looseness=-1

% To break this sequential dependency, recent works have explored multi-token prediction~\citep{gloeckle2024better, samragh2025your, liu2025tidar, chen2026dflash, kirchenbauer2026multi}, where multiple future tokens are proposed in a single drafting step. By avoiding token-by-token proposal generation, these approaches reduce sequential latency while still capturing locally predictable continuations, such as common phrases, syntactic patterns, or task-specific formats. Block diffusion language models~\citep{nie2025large, arriola2025block, wu2025fast, wu2025fast2} offer a natural instantiation of this idea: a diffusion drafter denoises a fixed block of mask tokens simultaneously, producing candidate distributions over future positions without left-to-right generation. This parallel structure is highly synergistic with speculative decoding, as demonstrated by TiDAR~\citep{liu2025tidar} and DFlash~\citep{chen2026dflash}, which report up to 2.5$\times$ speedups over AR drafting baselines.
% \looseness=-1

To break this sequential dependency, recent works explore multi-token prediction~\citep{gloeckle2024better, samragh2025your, liu2025tidar, chen2026dflash, kirchenbauer2026multi}, where multiple future tokens are proposed in a single drafting step. Block diffusion language models~\citep{nie2025large, arriola2025block, wu2025fast, wu2025fast2} provide a natural instantiation: a diffusion drafter denoises a fixed block of mask tokens simultaneously, producing candidate distributions over future positions without left-to-right generation. This parallel structure is well suited to speculative decoding, as shown by TiDAR~\citep{liu2025tidar} and DFlash~\citep{chen2026dflash}, which report up to 2.5$\times$ speedups over AR drafting baselines.
\looseness=-1

% Despite these latency advantages, parallel prediction introduces a critical challenge: because future positions are drafted before preceding tokens are finalized, the outputs are essentially sampled from position-wise marginal distributions rather than fully conditioned AR continuations. Consequently, greedy top-1 sampling often yields incoherent sequence under the target model's distribution~\citep{gu2017non, li2022diffusion, han2023ssd, wu2023ar}. Nevertheless, the target-preferred trajectory is highly likely to be included within the drafter's top-$K$ candidate tokens. In this work, motivated by this observation, we propose a \textbf{\textit{tree}-structured block-diffusion drafting} algorithm. Instead of committing to a single greedy path, we construct a prefix tree from the parallel candidate distributions, allowing the target model to verify multiple plausible trajectories simultaneously. This approach naturally inherits the core advantages of tree-based speculative decoding~\citep{miao2024specinfer, cai2024medusa, li2024eagle2fasterinferencelanguage}, improving verifier utilization and expected acceptance length.
% \looseness=-1

Despite these latency benefits, parallel prediction introduces a key challenge: future positions are drafted before preceding tokens are finalized, so the outputs correspond to position-wise marginals rather than fully conditioned AR continuations. Consequently, greedy top-1 sampling can yield incoherent sequences under the target model's distribution~\citep{gu2017non, li2022diffusion, han2023ssd, wu2023ar}. Yet the target-preferred trajectory often remains within the drafter's top-$K$ candidates. Motivated by this, we propose \textbf{\textit{tree}-structured block-diffusion drafting}: instead of committing to one greedy path, we construct a prefix tree from the parallel candidate distributions, allowing the target model to verify multiple plausible trajectories simultaneously. This inherits the core advantages of tree-based speculative decoding~\citep{miao2024specinfer, cai2024medusa, li2024eagle2fasterinferencelanguage}, improving verifier utilization and expected acceptance length.
\looseness=-1

% However, the practical efficiency of this tree-based approach heavily depends on its topology. Prior methods typically utilize heuristically predefined, static tree shapes that remain fixed regardless of the input context or hardware constraints~\citep{cai2024medusa, li2024eagle, park2025lantern++}. Thereby, we further propose \textbf{\textsc{BASTION}}, a \textbf{b}udget-\textbf{a}ware \textbf{s}peculative decoding with \textbf{t}ree-based diffus\textbf{ion} drafting. Unlike static approaches, \algo dynamically constructs a query-dependent tree topology to explicitly maximize the expected decoding speedup. Our method integrates three synergistic components: \textbf{(1)} an \textit{acceptance surrogate} that leverages the drafter's marginal distributions to estimate the expected accepted length via confidence scores of paths; \textbf{(2)} an \textit{online latency estimator} that predicts verification costs using a hardware-aware roofline model, calibrated by observed runtimes; and \textbf{(3)} an \textit{adaptive best-first expansion} mechanism that iteratively grows the tree and halts as soon as the marginal acceptance gain no longer outweighs the added verification cost.
% \looseness=-1

However, the efficiency of tree-based drafting depends critically on its topology. Prior methods typically rely on heuristically predefined, static tree shapes that remain fixed across inputs and hardware constraints~\citep{cai2024medusa, li2024eagle, park2025lantern++}. To address this, we propose \textbf{\textsc{BASTION}}, a \textbf{b}udget-\textbf{a}ware \textbf{s}peculative decoding method with \textbf{t}ree-based diffus\textbf{ion} drafting. Unlike static approaches, \algo dynamically constructs a query-dependent tree topology to maximize expected decoding speedup. It combines three components: \textbf{(1)} an \textit{acceptance surrogate} that estimates expected accepted length from path confidence scores under the drafter's marginal distributions; \textbf{(2)} an \textit{online latency estimator} that predicts verification cost with a hardware-aware roofline model calibrated by observed runtimes; and \textbf{(3)} an \textit{adaptive best-first expansion} mechanism that grows the tree until the marginal acceptance gain no longer outweighs the added verification cost.
\looseness=-1

% By systematically balancing draft quality against hardware capacity, \algo translates these theoretical optimizations into substantial empirical speedups. In \autoref{fig:main_result}, extensive evaluations across diverse domains---including math, code generation, and chat benchmarks---demonstrate that \algo achieves significant average speedups of $6.61\times$ (vs. standard decoding), $2.45\times$ (vs. EAGLE-3~\citep{li2025eagle}), and $1.39\times$ (vs. DFlash~\citep{chen2026dflash}). Furthermore, we comprehensively show that these significant performance gains are consistent and robust across multiple GPU architectures, highlighting the practical deployment viability of our dynamic tree-structured drafting paradigm.
% \looseness=-1

By balancing draft quality against hardware capacity, \algo delivers substantial empirical speedups. In \autoref{fig:main_result}, evaluations across math, code generation, and chat benchmarks show average speedups of $6.61\times$ over standard decoding, $2.45\times$ over EAGLE-3~\citep{li2025eagle}, and $1.39\times$ over DFlash~\citep{chen2026dflash}. These gains are consistent across multiple model and GPU architectures, highlighting the practical deployment viability of dynamic tree-structured drafting.
\looseness=-1

\vspace{-6pt}
\paragraph{Contributions.} In summary, our key contributions in this paper are as follows:
\vspace{-3pt}
\begin{itemize}[leftmargin=*, itemsep=2pt]
    \item \textbf{Tree-based block diffusion drafting:} 
    We introduce a best-first prefix-tree construction algorithm that uses drafter-logit path scores to build high-quality prefix-closed candidate trees from block-diffusion logits.
    % We use drafter-logit path scores to construct high-quality prefix-closed trees from block-diffusion logits, providing the fixed-budget inner optimizer used by our adaptive controller.
    \looseness=-1

    \item \textbf{Hardware-aware online budget controller:} To overcome the inefficiencies of fixed-size trees, we introduce a budget controller that dynamically determines the optimal tree size at each decoding step by balancing the expected acceptance gain against hardware-calibrated verification latency.
    % We show that fixed-budget tree construction is insufficient for end-to-end speedup, since larger trees improve average accepted length only until their marginal gain is outweighed by verification cost. We therefore introduce a train-free online controller that selects the tree size at each decoding step by combining drafter-logit-based acceptance estimation with calibrated, hardware-aware verifier-latency prediction.
    \looseness=-1

    \item \textbf{Extensive empirical validation:} 
    We validate \algo across target models, benchmarks, GPU architectures, and decoding temperatures, showing up to \textbf{$6.61\times$} speedup over autoregressive decoding and a \textbf{$38.87\%$} improvement over single-path block-diffusion drafting.
    % We comprehensively validate \algo across diverse target models, benchmarks, GPU architectures, and decoding temperatures, demonstrating consistent speedup improvements up to \textbf{$6.61\times$} over autoregressive and outperforms single-path block-diffusion drafting by \textbf{$38.87\%$}.
    \looseness=-1
    % We evaluate the method across target models, benchmarks, temperatures, and GPU platforms, showing consistent speedup improvements over single-path block-diffusion drafting \textcolor{red}{and ,fixed-budget tree-based}.
\end{itemize}

%% file: figures/overview.tex
\begin{figure}[!h]
  \centering
  % Leave room for the camera-ready author footnotes on the first page.
  \includegraphics[width=0.95\textwidth]{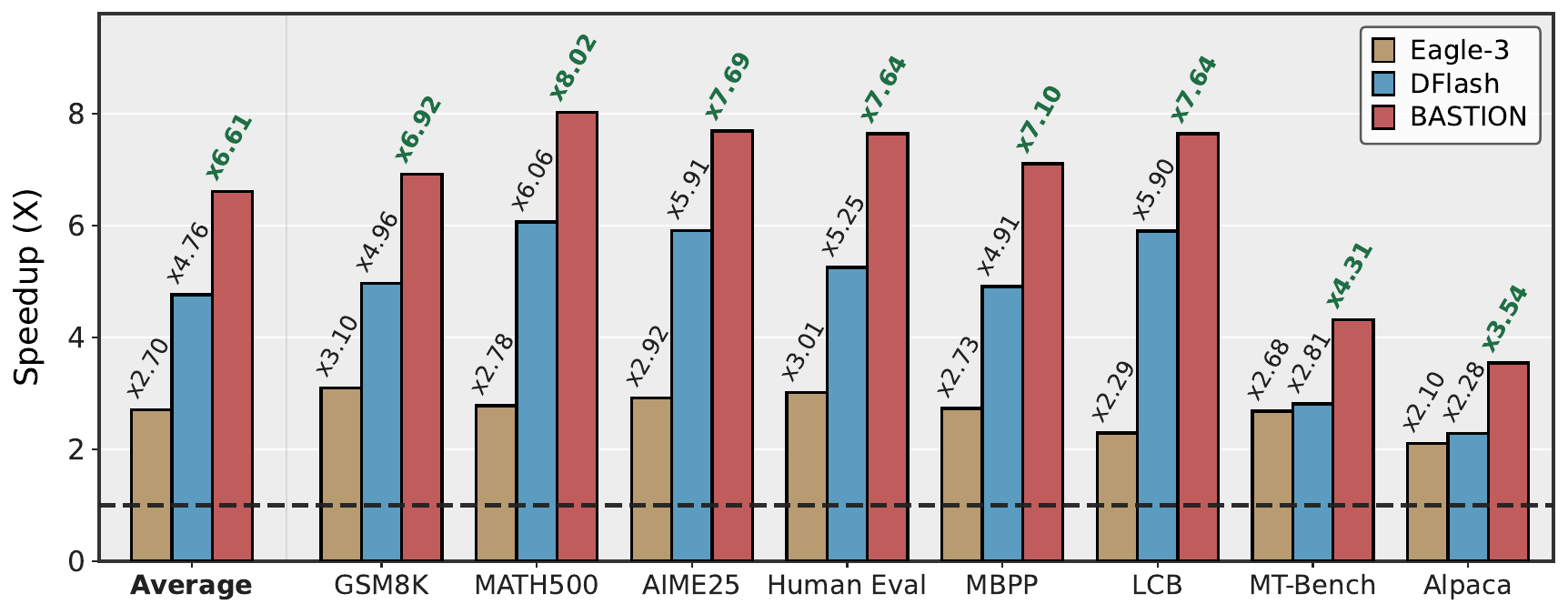}
  \vspace{-8pt}
  \caption{
  % \textbf{End-to-end speedup of speculative decoding methods on Qwen3-8B (A100, Temperature 0).} Per-dataset speedup over autoregressive decoding across the eight benchmarks. Ours consistently dominates DFlash and Eagle3. \sangmin{Add more details later}
  \textbf{\algo achieves a 6.61$\times$ average end-to-end speedup on Qwen3-8B.} \algo consistently outperforms speculative decoding baselines (EAGLE-3~\citep{li2025eagle} and DFlash~\citep{chen2026dflash}) across eight diverse benchmarks (three math, three code, and two chat datasets). The baseline performance (1$\times$) represents standard autoregressive decoding. Results are evaluated for a single sample using greedy decoding (i.e., temperature of 0) on an A100 GPU.
  }
  \label{fig:main_result}
\end{figure}

%% file: tex/03_prelim.tex
\section{Preliminary}
\label{sec:prelim}

\subsection{Speculative Decoding}

\paragraph{Mechanism overview.}

Speculative decoding accelerates generation by decoupling proposal generation from target-model verification~\citep{chen2023accelerating, leviathan2023fast, cai2024medusa, li2024eagle}. Instead of invoking the expensive target model sequentially, a lightweight draft model first proposes a sequence of future tokens. The target model then verifies these candidates in parallel with a single forward pass. 
Standard speculative sampling~\citep{chen2023accelerating, leviathan2023fast} probabilistically accepts tokens to theoretically match the target distribution. However, many recent approaches simplify this step by adopting exact match verification~\citep{chen2026dflash} or relaxed acceptance rules~\citep{cai2024medusa, bachmann2025judge}, bypassing the complex rejection sampling overhead while maintaining high throughput.
\looseness=-1

\vspace{-6pt}
\paragraph{Algorithm variants.}

Subsequent variants differ in their choice of drafters---ranging from model-free~\citep{saxena2023prompt, fu2024break} and autoregressive~\citep{leviathan2023fast, bae2023fast, elhoushi2024layerskip, li2025eagle} to diffusion-based approaches~\citep{samragh2025your, liu2025tidar, chen2026dflash}. While autoregressive drafters generate candidates through strict left-to-right sequential prediction, diffusion-based drafters instead produce draft spans by jointly denoising multiple future positions in parallel. Prior works have also explored various draft structures, such as sequential chains~\citep{chen2023accelerating, kim2023speculative, sadhukhan2024magicdec} and tree-based topologies~\citep{miao2024specinfer, cai2024medusa, chen_sequoia_2025}. Whereas sequential chains propose a single linear trajectory, tree-based methods simultaneously explore multiple diverging paths to increase the overall acceptance probability.
See \autoref{sec:related_work} for more details on related work.
\looseness=-1

\vspace{-6pt}
\paragraph{Efficiency analysis.}

Speculative decoding mitigates the memory-bandwidth bottleneck in LLM inference by using idle compute resources to verify multiple draft tokens in a single pass, trading extra computation for fewer memory accesses~\citep{chen2023accelerating, bae2023fast, sadhukhan2024magicdec}. Its practical speedup relative to standard decoding is defined as $\mathrm{Speedup} = \bar{A}\, \cdot T_{\mathrm{AR}} / (T_{\mathrm{draft}} + T_{\mathrm{verify}})$. This demonstrates that efficiency is maximized when the average number of accepted tokens per iteration ($\bar{A}$) is high, and the overhead times for generating drafts ($T_{\mathrm{draft}}$) and verifying them ($T_{\mathrm{verify}}$) are kept minimal compared to the latency of a standard single-token autoregressive step ($T_{\mathrm{AR}}$). 
\looseness=-1

\subsection{Block Diffusion Drafter}

\paragraph{Drafting algorithm.}

Unlike autoregressive drafting, which factorizes draft generation into sequential conditionals, one-step block diffusion drafting yields position-wise marginals under a shared masked context~\citep{liu2025tidar, li2025diffuspec, an2025pard, chen2026dflash}. A lightweight drafter—typically 5 to 8 layers—appends a sequence of \texttt{[MASK]} tokens to the context and denoises them simultaneously in a single forward pass, equivalent to masked token prediction~\citep{devlin2019bert, he2022masked}. Recent methods enhance drafter accuracy by injecting hidden representations from across the target model's layers directly into the drafter's key-value caches~\citep{liu2025tidar, chen2026dflash}. To extract a final draft trajectory from these parallel predictions, prior works employ greedy decoding, independently taking the top-1 prediction (i.e., \texttt{argmax}) at each masked position~\citep{liu2025tidar, chen2026dflash}. \looseness=-1

\vspace{-6pt}
\paragraph{Efficiency analysis.}

Recall that speculative speedup heavily depends on minimizing the drafting overhead $T_{\mathrm{draft}}$. For an autoregressive drafter, generating a draft block of $\gamma$ candidates requires $\gamma$ sequential forward passes, yielding a latency of $T_{\mathrm{draft}} = \gamma \cdot t_{\mathrm{step}}$, where $t_{\mathrm{step}}$ is the latency of a single forward pass. A block diffusion drafter circumvents this sequential bottleneck by processing the entire block in a single pass. Since LLM decoding is predominantly memory-bound, evaluating multiple tokens simultaneously incurs minimal computational overhead, effectively reducing the drafting latency to a near-constant $T_{\mathrm{draft}} \approx t_{\mathrm{step}}$. \looseness=-1

%% file: tex/04_method_new.tex
\section{\algo: Budget-Aware Tree Construction for Diffusion Drafting}
\label{sec:method}
 
In this section, we introduce our method \textbf{\algo}, a \textbf{b}udget-\textbf{a}ware \textbf{s}peculative decoding framework with \textbf{t}ree-based diffus\textbf{ion} drafting. This section
formalizes three components: (i) a problem formulation that exploits the
parallel nature of block-diffusion drafting (\S\ref{sec:method:formulation}),
(ii) a path-confidence surrogate for acceptance estimation (\S\ref{sec:method:surrogate},\S\ref{sec:method:fixed}), and (iii) an online
controller for budget optimization (\S\ref{sec:method:adaptive}).
 
% --------------------------------------------------------------------------
\subsection{Problem Formulation: Tree-based Block-Diffusion Drafting}
\label{sec:method:formulation}
 
A block-diffusion drafter produces position-wise marginal distributions
$\{q_k(\cdot)\}_{k=1}^{\gamma}$ in a single parallel forward pass at constant
cost $T_{\mathrm{draft}} = t_{\mathrm{parallel}}$, whereas an \textit{autoregressive} (AR)
drafter incurs $T_{\mathrm{draft}} = \gamma\cdot t_{\mathrm{step}}$. Two
consequences shape our design:
\begin{itemize}
\item[(P1)] \textbf{Topology decoupling.} $T_{\mathrm{draft}}$ is invariant to
  the candidate-set structure---chains, wide trees, and deep trees incur
  identical drafting cost.
\item[(P2)] \textbf{Marginal independence.} Each $q_k$ is conditioned on the
  validated context alone, not on drafted tokens at positions
  $\{1,\dots,k-1\}$, so candidates across positions can be combined freely.
\end{itemize}

These properties motivate organizing the candidates as a \emph{prefix tree}
$\mathcal{T}$: the root $r$ is the validated bonus token, and each non-root
node $i$ at depth $d(i)\in\{1,\dots,\gamma\}$ carries a candidate token $x_i$
whose root-to-node path
$\mathrm{path}(i)=(x_{i_1},\dots,x_{i_{d(i)}})$ receives the \emph{path score}
\begin{equation}
\rho(i) \;=\; \prod_{k=1}^{d(i)} q_k(x_{i_k}),\qquad \rho(r) = 1.
\label{eq:path-score}
\end{equation}
The target model verifies all nodes in one parallel forward pass. Starting
from the root, we follow the child matching the target's next token at each
visited prefix, sampled for stochastic decoding or selected by argmax for
greedy decoding, and commit this path until no child matches. Counting the always-committed root,
let $A(\mathcal{T})$ be the resulting committed length and define the
\emph{verification budget} $N = |\mathcal{T}|$. The end-to-end speedup over
AR decoding is then
\begin{equation}
\mathrm{Speedup}(\mathcal{T})
\;=\;
\frac{\mathbb{E}[A(\mathcal{T})]\cdot L_{\mathrm{AR}}}
     {T_{\mathrm{draft}} + T_{\mathrm{aux}}(\mathcal{T}) + T_{\mathrm{verify}}(\mathcal{T})},
\label{eq:speedup}
\end{equation}
where $L_{\mathrm{AR}}$ is one AR target step, $T_{\mathrm{aux}}(\mathcal{T})$
covers tree construction and scheduling, and $T_{\mathrm{verify}}(\mathcal{T})$
is the parallel verification cost. Our goal is to choose $\mathcal{T}$---both
its shape and size $N$---to maximize \Eqref{eq:speedup} from the drafter's
logits alone.

% --------------------------------------------------------------------------
\subsection{Expected Committed Length Surrogate}
\label{sec:method:surrogate}
Maximizing \Eqref{eq:speedup} before verification is intractable: the realized
committed length $A(\mathcal{T})$ depends on the target's outputs, which are
unavailable until verification completes. We therefore approximate
$\mathbb{E}[A(\mathcal{T})]$ via a \emph{drafter-side surrogate} obtained from
a self-verification thought experiment.
 
\paragraph{Self-verification model.}
Replace the target with the drafter itself as the verifier: the drafter draws
a single sample $X = (X_1,\dots,X_\gamma)$ with $X_k \sim q_k(\cdot)$
independently across positions, and a node $i\in\mathcal{T}$ is committed iff
$X_{1:d(i)} = \mathrm{path}(i)$. By property (P2), this product distribution
is \emph{precisely} the drafter's joint distribution over $\gamma$-length
continuations, so this model is the natural drafter analogue of target
verification---not an approximation of the drafter, only of the target. Let
$A_{\mathrm{self}}(\mathcal{T};X)$ denote the resulting committed length
(including the always-committed root).
 
\begin{lemma}[Committed length equals covered count]
\label{lem:cover-count}
Under the self-verification model,
\[
A_{\mathrm{self}}(\mathcal{T};X)
\;=\;
\#\!\left\{\, i \in \mathcal{T} \,:\, X_{1:d(i)} = \mathrm{path}(i) \,\right\}.
\]
The set of committed nodes forms a contiguous root-to-node chain in
$\mathcal{T}$.
\end{lemma}
 
The proof (Appendix~\ref{app:proofs:b1}) follows from two structural facts:
prefix-closure of $\mathcal{T}$ forces the covered set to be ancestor-closed,
and the tree structure permits at most one covered node per depth; hence the
covered set is a single chain whose length equals its cardinality.
 
\begin{proposition}[Path-Sum Surrogate]
\label{prop:path-sum}
Taking expectation over the drafter's sample $X$,
\begin{equation}
\widehat{A}(\mathcal{T})
\;\triangleq\;
\mathbb{E}_X\!\left[A_{\mathrm{self}}(\mathcal{T};X)\right]
\;=\;
\sum_{i\in\mathcal{T}} \rho(i).
\label{eq:surrogate}
\end{equation}
\end{proposition}
 
The proof (Appendix~\ref{app:proofs:b1}) applies linearity of expectation to
Lemma~\ref{lem:cover-count}, using
$\mathbb{P}(X_{1:d(i)}=\mathrm{path}(i)) = \rho(i)$ by position-wise
independence and the convention $\rho(r)=1$ for the trivially covered root.
 
\paragraph{Why $\widehat{A}$ proxies the target-side expectation.}
\Eqref{eq:surrogate} is \emph{exact} only under self-verification,
where the drafter acts as both proposer and verifier. Its use as a proxy for
the target-verified $\mathbb{E}[A(\mathcal{T})]$ instead relies on
drafter--target alignment, the foundational premise of speculative decoding:
when the drafter's high-probability continuations coincide with the target's,
the drafter-mass covered by $\mathcal{T}$ closely tracks the realized accepted
length. Appendix~\ref{app:top1_path_score_validation}
(Figure~\ref{fig:mean_al_correlate}) empirically validates this proxy with
Pearson correlation $\geq 0.79$ across all target/draft pairs.

\input{figures/acceptance_latency_tradeoff}
 
% --------------------------------------------------------------------------
\subsection{Optimal Tree Construction via Best-First Expansion}
\label{sec:method:fixed}
 
With the surrogate in hand, the fixed-budget tree-construction problem
becomes: \emph{given $N$, find the prefix-closed tree
$\mathcal{T}\subseteq\mathcal{V}$ of size $N$ that maximizes
$\widehat{A}(\mathcal{T})$,} where $\mathcal{V}$ denotes the candidate lattice
induced by retaining the top-$K$ tokens at each position. Beam-search-style
methods, which fix a $(\text{width}\times\text{depth})$ allocation, are
mismatched here because they ignore the global ordering of path confidences
across depths.
 
A key structural property is \emph{path monotonicity}: for any non-root $i$
with parent $\pi(i)$,
\begin{equation}
\rho(i) \;=\; \rho(\pi(i))\cdot q_{d(i)}(x_i)\;\le\;\rho(\pi(i)),
\label{eq:path-monotone}
\end{equation}
because $q_{d(i)}(x_i)\in[0,1]$. This monotonicity, combined with
prefix-closure, makes a greedy procedure provably optimal.
 
\begin{proposition}[Optimality of Best-First Expansion]
\label{prop:bestfirst}
Iteratively adding the valid node with the largest $\rho$ yields nested trees $\mathcal{T}_1\subset\mathcal{T}_2\subset\cdots\subset\mathcal{T}_{N_{\max}}$ where each $\mathcal{T}_N$ maximizes $\widehat{A}(\mathcal{T})$ among all size-$N$ prefix-closed trees drawn from $\mathcal{V}$.
\end{proposition}
 
The proof (Appendix~\ref{app:proofs:b2}) demonstrates that best-first enumerates nodes in a globally non-increasing order of $\rho$. Consequently, the marginal gain $\Delta\widehat{A}(N) :=
\widehat{A}(\mathcal{T}_{N+1}) - \widehat{A}(\mathcal{T}_N) = \rho(i_{N+1})$ is non-increasing:
\begin{equation}
\Delta\widehat{A}(N+1) \;\le\; \Delta\widehat{A}(N)
\quad\text{for all } N,
\label{eq:concave-A}
\end{equation}
making $\widehat{A}$ is \emph{concave} in the budget $N$. As shown in Figure~\ref{fig:latency_accept_tradeoff} (left), this concavity yields diminishing returns, directly motivating the need for an adaptive budget strategy. 
 
% --------------------------------------------------------------------------
\input{figures/adaptive_tree_diagram}

\subsection{Online Controller for Budget Optimization}
\label{sec:method:adaptive}
 
A fixed-budget construction maximizes $\widehat{A}(\mathcal{T})$ for a
\emph{given} $N$ but is silent on which $N$ to use. No single $N$ is
universally optimal: the marginal acceptance gain depends on the local draft
distribution, while the verification cost depends on context length, model
architecture, and hardware. As illustrated in Figure~\ref{fig:latency_accept_tradeoff} (right), verification costs $T_{\text{verify}}$ scale with $N=|\mathcal T|$ and eventually dominate the cycle. To formalize this trade-off, we maximize the \emph{estimated speedup at budget $N$}:
\begin{equation}
S(N) \;=\;
\frac{\widehat{A}(\mathcal{T}_N)\cdot L_{\mathrm{AR}}}{C(N)},
\qquad
\begin{aligned}
C(N) &\;\triangleq\;
T_{\mathrm{draft}} + T_{\mathrm{aux}}(N) + T_{\mathrm{verify}}(N) \\
&\approx T_{\mathrm{draft}} + T_{\mathrm{aux}} + T_{\mathrm{verify}}(N)
\end{aligned}
\label{eq:speedup-N}
\end{equation}

Figure~\ref{fig:latency_accept_tradeoff} (right) shows that $T_{\mathrm{aux}}(N)$ is negligible relative to verification, so we approximate it by a constant $T_{\mathrm{aux}}$. We predict $T_{\mathrm{verify}}(N)$ with a hardware-aware roofline model (Appendix~\ref{app:sec:cost_model}). For dense Transformers, this yields a convex verification cost and hence a convex approximated cycle cost $C(N)$; the appendix discusses architecture-specific MoE extensions and their convexity limits. We apply linear calibration to account for kernel fusion, cache effects, scheduling, and tree-attention overhead.

\begin{proposition}[Unimodality of Estimated Speedup]
\label{prop:unimodal}
Suppose $\widehat{A}(N)$ is concave and non-decreasing, and $C(N)$ is convex
and strictly positive. Then $S(N)$ is unimodal on
$\{1,\dots,N_{\max}\}$. In particular, the smallest $N$ for which $S(N{+}1) < S(N)$ is a global maximizer.
\end{proposition}
 
The proof is provided in Appendix~\ref{app:proofs:b3}. Propositions~\ref{prop:bestfirst} and~\ref{prop:unimodal} combine into a
single online procedure as illustrated in \autoref{fig:tree_expansion}. Best-first expansion produces a nested chain
$\mathcal{T}_1\subset\mathcal{T}_2\subset\cdots$, with each step adding one
node $i_{N+1}$ and updating the surrogate incrementally as
$\widehat{A}(\mathcal{T}_{N+1}) = \widehat{A}(\mathcal{T}_N) + \rho(i_{N+1})$.
We pair this with a hardware-calibrated estimate $\widehat{C}(N)$ of cycle
latency and track
$\widehat{S}(N) = \widehat{A}(\mathcal{T}_N)\,L_{\mathrm{AR}} / \widehat{C}(N)$.
The controller halts at the first $N$ where $\widehat{S}(N{+}1) < \widehat{S}(N)$
and returns $\mathcal{T}_N$; Proposition~\ref{prop:unimodal} guarantees this
stopping rule when its concavity and convexity premises hold. The
per-step overhead is dominated by best-first heap operations and is negligible
compared to the target forward pass. Pseudocode is given in
Algorithm~\ref{alg:adaptive_tree_search}.

%% file: figures/acceptance_latency_tradeoff.tex
\begin{figure*}[t]
    \centering
    \begin{subfigure}[t]{0.43\textwidth}
        \includegraphics[width=\textwidth]{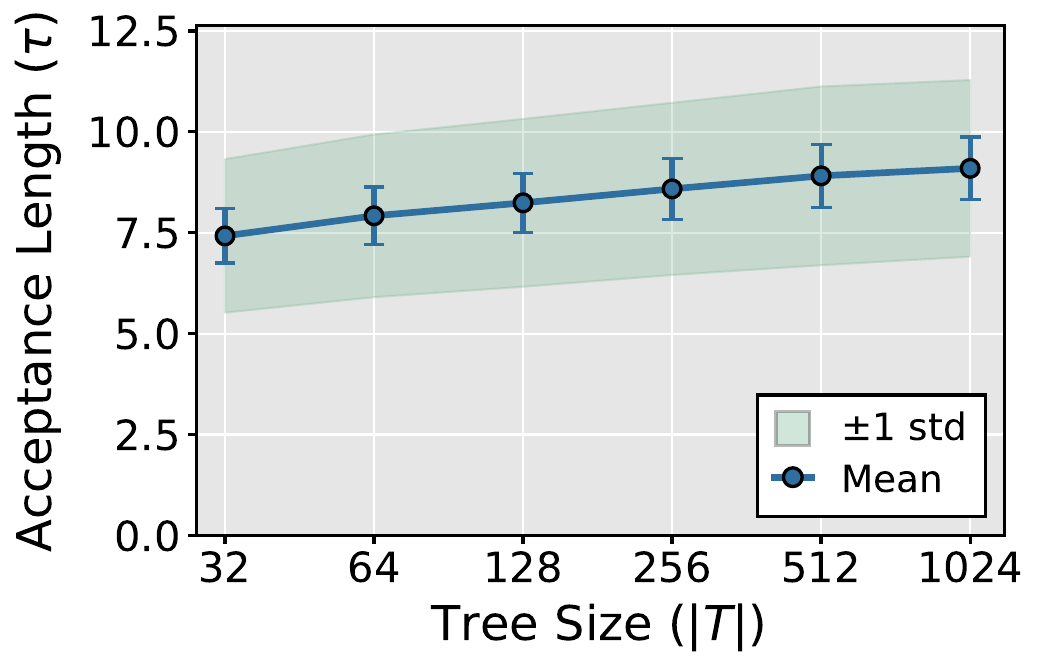}
    \end{subfigure}
    \hspace{20pt}
    \centering
    \begin{subfigure}[t]{0.43\textwidth}
        \includegraphics[width=\textwidth]{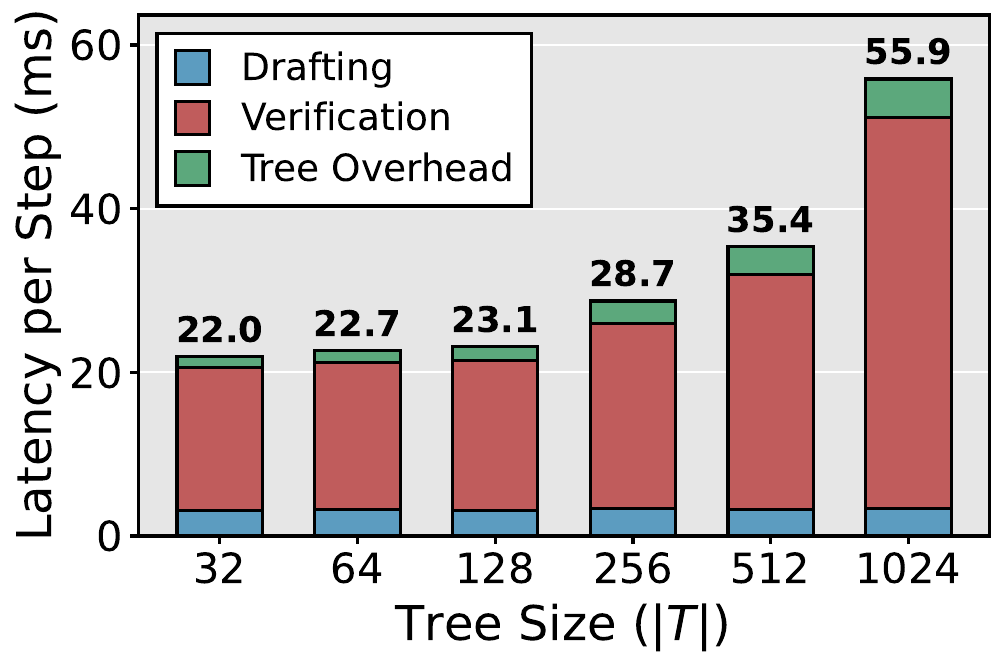}
    \end{subfigure}
    \caption{\textbf{Acceptance--latency trade-off across tree sizes.}
    \emph{Left:} acceptance length $\tau$ grows with tree size $|\mathcal{T}|$ but saturates beyond a few hundred nodes, reflecting diminishing marginal gains.
    \emph{Right:} per-step latency breakdown---drafting cost is constant, while $T_{\mathrm{aux}}$ and $T_{\mathrm{verify}}$ grow with $|\mathcal{T}|$, with $T_{\mathrm{verify}}$ dominating at large budgets ($22.0$\,ms at $|\mathcal{T}|{=}32$ rising to $55.9$\,ms at $|\mathcal{T}|{=}1024$).}
    \label{fig:latency_accept_tradeoff}
\end{figure*}

%% file: figures/adaptive_tree_diagram.tex
% \begin{figure}[!tbp]
%   \centering
%   \includegraphics[width=\textwidth]{assets/tree_expansion_combined.pdf}
%   \caption{\textbf{Adaptive tree construction from block-diffusion logits.}
%   \textbf{(a)} The drafter provides top-$K$ candidates for multiple future
%   positions in one forward pass, inducing an implicit lattice of candidate
%   prefixes. \textbf{(b)} Best-first expansion adds nodes in descending path
%   probability $\rho(i)$ and evaluates the estimated speedup $\widehat S_t(N)$
%   after each intermediate budget. The controller returns the tree with the
% f  largest estimated speedup instead of using a fixed depth or width.}
%   \label{fig:tree_expansion}
% \end{figure}

\begin{figure*}[t]
    \centering
    \begin{subfigure}[t]{0.47\textwidth}
        \includegraphics[width=\textwidth]{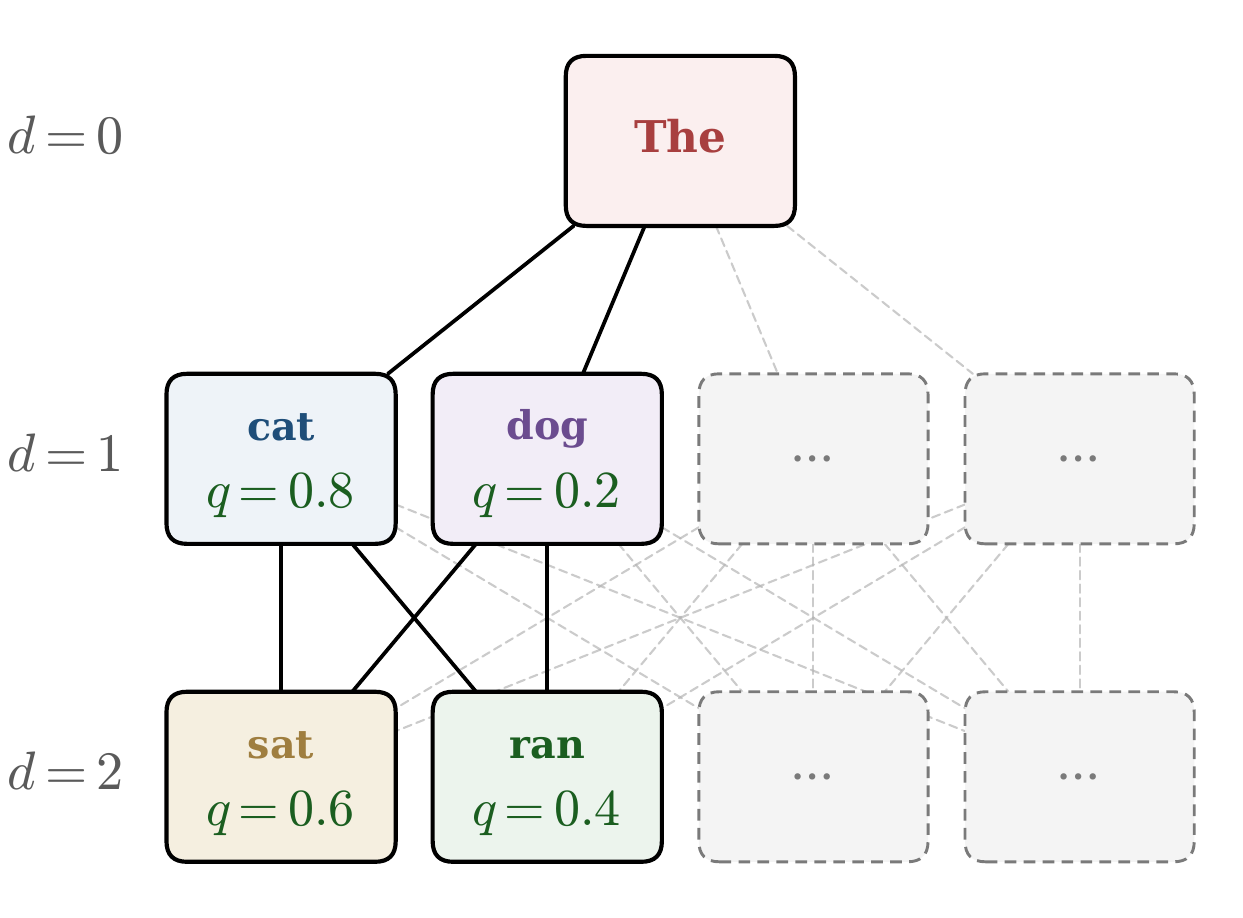}
        \subcaption{Top-$K$ logits per position}
    \end{subfigure}
    \hfill
    \centering
    \begin{subfigure}[t]{0.52\textwidth}
        \includegraphics[width=\textwidth]{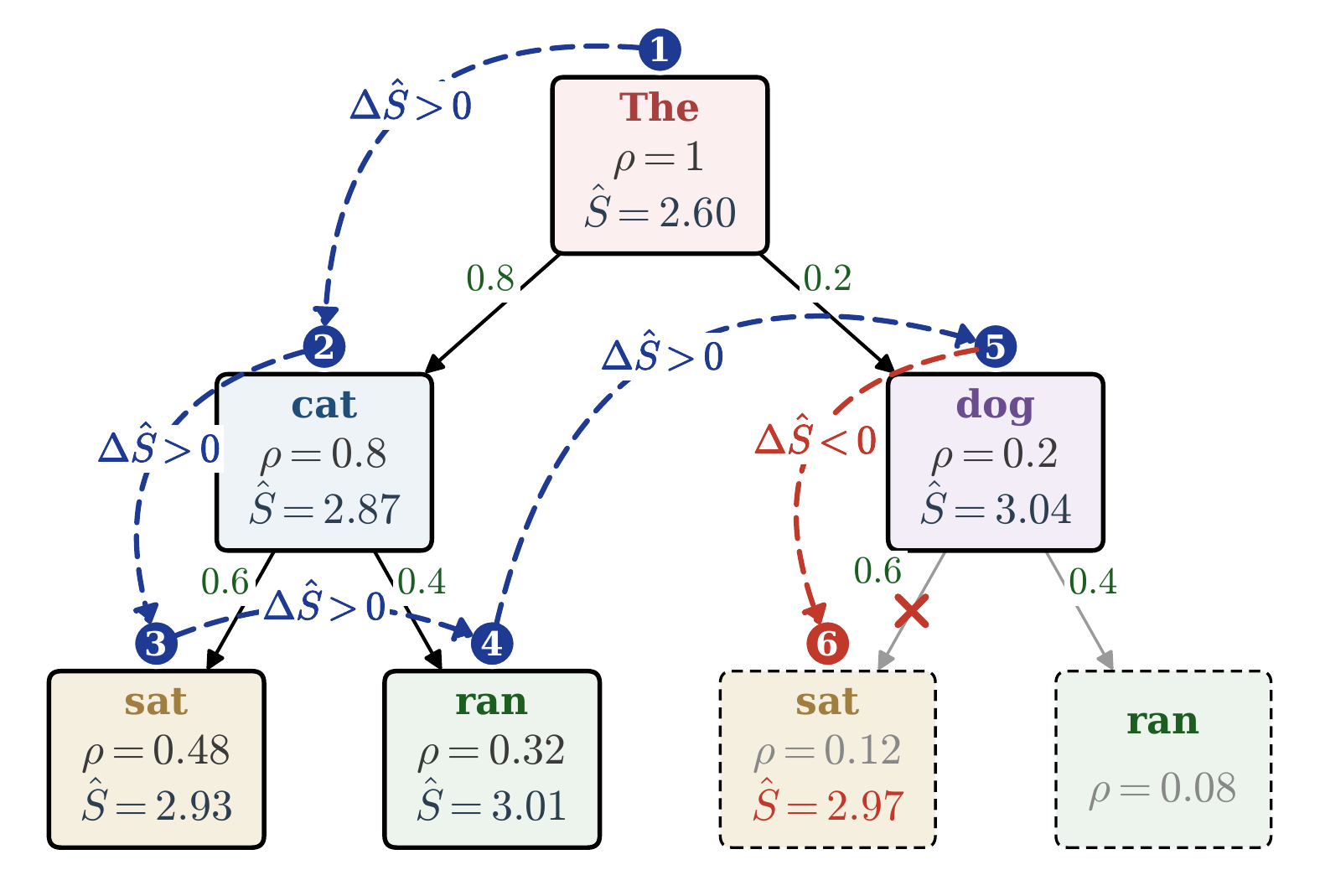}
        \subcaption{Adaptive tree construction}
    \end{subfigure}
      \caption{\textbf{Adaptive tree construction from block-diffusion logits.}
      \textbf{(a)} The drafter provides top-$K$ candidates for multiple future
      positions in one forward pass, inducing an implicit lattice of candidate
      prefixes. \textbf{(b)} Best-first expansion adds nodes in descending path
      probability $\rho(i)$ and evaluates the estimated speedup $\widehat S_t(N)$
      after each intermediate budget. The controller returns the tree with the largest estimated speedup instead of using a fixed depth or width.}
      \label{fig:tree_expansion}
\end{figure*}

%% file: tex/05_experiments.tex
\section{Experiments}
\label{sec:experiments}

\subsection{Experimental Settings}
\label{experimental_settings}

We evaluate \algo using Qwen3 (4B/8B) and Llama-3.1-8B-Instruct paired with DFlash block-diffusion drafters. We compare our method against DFlash and EAGLE-3 across four primary domains: mathematical reasoning, code generation, instruction following, and long-context understanding. Detailed experimental settings, including comprehensive dataset lists, metrics, and hardware configurations, are deferred to Appendix~\ref{app:sec:detail_exp_setup}.

\input{tables/main_experiment_a100}

\input{figures/addition_gpu_model_results}

\subsection{Main Results}
\label{subsec:main_results}

\paragraph{End-to-end speedup.}
\autoref{tab:main_res} reports comparative results across two Qwen3 target models, two decoding temperatures, and eight benchmarks on NVIDIA A100. \algo improves speedup over both DFlash and EAGLE-3 in all settings, with the largest gains on math and code benchmarks. Compared with DFlash, the improvement comes from using the drafter's full position-wise distribution to construct a tree rather than verifying only a single top-1 continuation. Compared with EAGLE-3, the gain reflects the lower draft latency of block-diffusion drafting together with adaptive tree-budget selection.
\looseness=-1

\vspace{-6pt}
\paragraph{Average acceptance length.}
Across benchmarks, \algo obtains substantially larger AAL than DFlash, indicating that tree-structured verification accepts longer continuations than single-path drafting. This confirms that the acceptance surrogate and best-first expansion allocate verification nodes to useful candidate paths. The speedup gain is not determined by AAL alone, however: the selected budget must also avoid excessive verifier latency, which motivates the budget-sweep analysis in Section~\ref{subsec:fixed_budget_sweep}.
\looseness=-1

\vspace{-6pt}
% \paragraph{Generalization across GPUs and model families.}
% The same trends persist beyond A100: \algo{} retains its lead on the A6000 and RTX PRO 6000 Blackwell across all three target models (Qwen3-4B, Qwen3-8B, and Llama-3.1-8B-Instruct; \autoref{fig:additional_results}), and on H100 with Qwen3 backbones (Appendix~\ref{app:subsec:h100}), indicating that the improvements are not tied to a specific GPU generation or model family.
% \looseness=-1
\paragraph{Generalization across GPUs and model families.}
Beyond A100, \algo{} retains its lead on A6000 and RTX PRO 6000 Blackwell across Qwen3-4B/8B and Llama-3.1-8B-Instruct (\autoref{fig:additional_results}), and on H100 (Appendix~\ref{app:subsec:h100}). Appendix~\ref{app:subsec:moe} extends the evaluation to the MoE target Qwen3-Coder-30B-A3B-Instruct on coding benchmarks.
\looseness=-1

% \paragraph{Static and online calibration offer different deployment trade-offs.} 
% \textsc{BASTION}-Static uses an offline-calibrated latency model and provides a deterministic default when model--GPU calibration data is available. \textsc{BASTION}-EMA updates the latency correction online from observed runtimes and is useful when offline calibration is unavailable or the runtime environment shifts. Both variants consistently outperform the fixed baselines, suggesting that the main benefit comes from adaptive budget selection rather than a particular calibration protocol.

\subsection{Tree Expansion Comparison}
\label{subsec:tree_expansion_ablation}
\input{tables/best_first_vs_beam_search}

\paragraph{Tree topology comparison.}
We ablate the tree expansion rule under a matched verification budget $N$, comparing the best-first expansion of \algo against standard beam search. Unlike beam search, which uniformly retains the top $w$ extensions depth-by-depth to form a rigid $(w{\times}d)$ topology, best-first dynamically expands the highest-scoring open nodes globally based on cumulative path scores $\rho(i)$ (\Eqref{eq:path-score}). \autoref{fig:tree_style_comparison}(a) illustrates this for $N{=}17$: beam search yields a fixed rectangle, whereas best-first flexibly deepens high-confidence paths. By Proposition~\ref{prop:bestfirst}, best-first produces the optimal prefix-closed tree under the surrogate $\widehat{A}$; this ablation evaluates whether this theoretical optimality translates to wall-clock speedups over beam search. Additionally, \autoref{fig:tree_style_comparison}(b) presents a non-budget-matched single-path DFlash baseline (greedy, block size $16$) to isolate the inherent benefits of utilizing a tree structure over a no-tree reference.

% We ablate the tree expansion rule under a matched verification budget $N$, comparing best-first expansion (the construction used by \algo) against beam search---the standard topology in tree-based speculative decoding. Beam search proceeds depth by depth: at each level it retains the top $w$ extensions of every surviving prefix, yielding a rigid $(w{\times}d)$ rectangle independent of where draft confidence concentrates. Best-first expansion instead ranks all prefix-completable candidates globally by their cumulative path score $\rho(i)$ (\Eqref{eq:path-score}) and adds the highest-scoring open node at each step; the resulting topology is shaped by the drafter's distribution rather than a fixed schedule. \autoref{fig:tree_style_comparison}(a) makes this concrete at $N{=}17$: beam search produces a uniform $4{\times}4$ rectangle plus root, whereas best-first deepens a small number of high-confidence prefixes and prunes others entirely. By Proposition~\ref{prop:bestfirst}, the best-first tree is the optimal prefix-closed tree of size $N$ under the path-sum surrogate $\widehat{A}$; this ablation tests whether that surrogate-level optimality translates to wall-clock speedup against the standard beam-search baseline. We additionally report the single-path DFlash baseline (greedy, block size $16$) in \autoref{fig:tree_style_comparison}(b) as a no-tree reference: this row uses a smaller per-step budget than the two tree variants and is therefore \emph{not} budget-matched, but it isolates the contribution of tree structure itself from the choice of expansion rule.

\vspace{-6pt}
% \paragraph{Best-first expansion outperforms beam search under matched budgets.}
% \autoref{fig:tree_style_comparison}(b) shows that best-first expansion improves both mean speedup and average acceptance length $\tau$ under a matched per-step budget (beam: $w{=}4, d{=}15$; best-first: $N{=}61$). On Qwen3-4B and Qwen3-8B, best-first attains $+7.0\%$ and $+6.1\%$ mean speedup respectively, with higher $\tau$ across all benchmarks. Both tree variants substantially outperform the single-path greedy baseline even though greedy uses a smaller budget, but the budget-matched comparison isolates topology as the source of best-first's gain over beam search. Concentrating tree expansion on high-scoring prefixes, rather than spreading it uniformly across depths, yields more efficient speculative decoding with block-diffusion drafters.

\paragraph{Best-first expansion outperforms beam search under matched budgets.}
\autoref{fig:tree_style_comparison}(b) shows best-first expansion improves mean speedup and average acceptance length $\tau$ under a matched budget (beam: $w{=}4, d{=}15$; best-first: $N{=}61$). On Qwen3-4B and Qwen3-8B, best-first yields $+7.0\%$ and $+6.1\%$ speedups respectively, with consistently higher $\tau$. While both tree variants outperform the greedy baseline, this comparison isolates topology as the source of best-first's advantage. Concentrating expansion on high-scoring prefixes, rather than uniformly across depths, enables more efficient block-diffusion drafting.

\subsection{Multi-Batch Results}
\label{subsec:sglang}

The preceding experiments use batch size~1. To test whether the adaptive
controller still pays off once requests are batched, we evaluate \algo
end-to-end inside SGLang~\citep{zheng2024sglang} at request concurrency
$C\in\{1,4,8,16,32,64\}$, reusing one calibration throughout.

\input{tables/sglang_throughput}

\autoref{tab:sglang_throughput} compares \algo against DFlash and against its own
budget frozen at $N\in\{16,32,64\}$. Untuned, \algo reaches $98.6\%$ of the best
fixed budget on average and is the fastest method at $C{=}16$ and $C{=}64$. The
highest-acceptance configuration is the slowest: $N{=}64$ attains $\tau=9.22$ yet
delivers only $573$ tokens/s, roughly half the throughput of $N{=}16$. Acceptance
alone is therefore not the serving objective. Balancing it against verification
cost is exactly what the latency-aware controller of Section~\ref{sec:method}
does.

\input{tables/sglang_per_step_latency}

\autoref{tab:sglang_per_step} breaks this cost down at an effective batch of
about $36$, comparing DFlash's single-path (chain) verification with a fixed
tree of $N{=}16$. The tree is more expensive in both stages, even though it
verifies no more tokens per request than the chain. \emph{Preparation} rises
by $49.3\%$ because the tree builder needs normalized top-$K$ probabilities
over the full vocabulary at every draft position, followed by a best-first
expansion for each request, whereas the chain only needs the top-1 token at
each position. \emph{Verification} rises by $15.1\%$ because each tree node may
attend only to its ancestors, which requires a custom attention mask for each
request instead of the causal mask that attention kernels are optimized for.
At batch size~1, both overheads are small next to the fixed cost of reading
the model weights. Batching amortizes that fixed cost across requests, while
the tree overheads grow with the number of requests, so each extra node costs
more as concurrency rises. This is why the larger fixed trees in
\autoref{tab:sglang_throughput} lose their lead over DFlash as $C$ grows.
\algo captures this rising cost in its latency profile and contracts the
budget accordingly.

%% file: tables/main_experiment_a100.tex
\begin{table*}[!t]
\caption{\textbf{Comparison of speedup and AAL ($\tau$) on NVIDIA A100 gpu.} The experiments are conducted on Qwen3-4B and Qwen3-8B at Temperature $\in \{0, 1\}$. For the baselines, the tree budget size of EAGLE-3 is set to 60, and the block size of DFlash is set to 16.}
\label{tab:main_res}
\centering
\small
\setlength{\tabcolsep}{1pt}
\resizebox{\textwidth}{!}{
\begin{tabular}{l | c | *{3}{c c} | *{3}{c c} |*{2}{c c} | *{1}{c c}}
\toprule
\multirow{2.5}{*}{\textbf{Model}} & \multirow{2.5}{*}{\textbf{Method}} & \multicolumn{2}{c}{\textbf{GSM8K}} & \multicolumn{2}{c}{\textbf{MATH500}} & \multicolumn{2}{c}{\textbf{AIME25}} & \multicolumn{2}{|c}{\textbf{HumanEval}} & \multicolumn{2}{c}{\textbf{MBPP}} & \multicolumn{2}{c}{\textbf{LCB}} & \multicolumn{2}{|c}{\textbf{MT-Bench}} & \multicolumn{2}{c}{\textbf{Alpaca}} & \multicolumn{2}{|c}{\textbf{Average}} \\
\cmidrule(lr){3-4} \cmidrule(lr){5-6} \cmidrule(lr){7-8} \cmidrule(lr){9-10} \cmidrule(lr){11-12} \cmidrule(lr){13-14} \cmidrule(lr){15-16} \cmidrule(lr){17-18} \cmidrule(lr){19-20}
& & Speedup & $\tau$ & Speedup & $\tau$ & Speedup & $\tau$ & Speedup & $\tau$ & Speedup & $\tau$ & Speedup & $\tau$ & Speedup & $\tau$ & Speedup & $\tau$ & Speedup & $\tau$ \\
\midrule
\multicolumn{20}{c}{\textcolor{gray!90}{\textbf{\textit{Greedy Decoding (Temperature = 0)}}}} \\
\midrule
& EAGLE-3 & $3.11\times$ & 2.77 & $2.84\times$ & 2.53 & $2.92\times$ & 2.52 & $2.93\times$ & 2.57 & $2.77\times$ & 2.48 & $2.66\times$ & 2.10 & $2.69\times$ & 2.48 & $2.45\times$ & 2.15 & $2.80\times$ & 2.45 \\
& DFlash & $5.23\times$ & 6.50 & $6.22\times$ & 8.12 & $6.16\times$ & 7.48 & $5.34\times$ & 6.66 & $5.09\times$ & 6.31 & $5.71\times$ & 7.23 & $2.74\times$ & 4.63 & $2.12\times$ & 2.90 & $4.83\times$ & 6.23 \\
\rowcolor{blue!10} \cellcolor{white} \multirow{-3}{*}{Q3-4B} & \textbf{\algo} & $\mathbf{7.44\times}$ & \textbf{9.07} & $\mathbf{8.57\times}$ & \textbf{10.60} & $\mathbf{8.23\times}$ & \textbf{9.70} & $\mathbf{7.91\times}$ & \textbf{9.38} & $\mathbf{7.41\times}$ & \textbf{8.99} & $\mathbf{7.81\times}$ & \textbf{9.48} & $\mathbf{4.53\times}$ & \textbf{6.58} & $\mathbf{3.53\times}$ & \textbf{4.64} & $\mathbf{6.93\times}$ & \textbf{8.56} \\
\cmidrule{1-20}
& EAGLE-3 & $3.10\times$ & 2.74 & $2.78\times$ & 2.51 & $2.92\times$ & 2.45 & $3.01\times$ & 2.69 & $2.73\times$ & 2.24 & $2.29\times$ & 1.91 & $2.68\times$ & 2.24 & $2.10\times$ & 1.77 & $2.70\times$ & 2.32 \\
& DFlash & $4.96\times$ & 6.51 & $6.06\times$ & 8.25 & $5.91\times$ & 7.20 & $5.25\times$ & 6.64 & $4.91\times$ & 6.04 & $5.90\times$ & 7.64 & $2.81\times$ & 4.46 & $2.28\times$ & 3.15 & $4.76\times$ & 6.24 \\
\rowcolor{blue!10} \cellcolor{white} \multirow{-3}{*}{Q3-8B} & \textbf{\algo} & $\mathbf{6.92\times}$ & \textbf{8.85} & $\mathbf{8.02\times}$ & \textbf{10.53} & $\mathbf{7.69\times}$ & \textbf{9.46} & $\mathbf{7.64\times}$ & \textbf{9.44} & $\mathbf{7.10\times}$ & \textbf{8.65} & $\mathbf{7.64\times}$ & \textbf{9.91} & $\mathbf{4.31\times}$ & \textbf{6.28} & $\mathbf{3.54\times}$ & \textbf{4.68} & $\mathbf{6.61\times}$ & \textbf{8.48} \\
\midrule
\multicolumn{20}{c}{\textcolor{gray!90}{\textbf{\textit{Stochastic Decoding (Temperature = 1)}}}} \\
\midrule
& EAGLE-3 & $3.03\times$ & 2.75 & $2.69\times$ & 2.44 & $2.61\times$ & 2.21 & $2.61\times$ & 2.44 & $2.70\times$ & 2.41 & $2.38\times$ & 2.08 & $2.59\times$ & 2.37 & $2.41\times$ & 2.16 & $2.63\times$ & 2.36 \\
& DFlash & $4.58\times$ & 6.04 & $4.93\times$ & 6.77 & $3.51\times$ & 4.75 & $4.67\times$ & 6.08 & $4.42\times$ & 5.90 & $4.92\times$ & 6.64 & $2.56\times$ & 4.28 & $2.04\times$ & 2.81 & $3.95\times$ & 5.41 \\
\rowcolor{blue!10} \cellcolor{white} \multirow{-3}{*}{Q3-4B} & \textbf{\algo} & $\mathbf{6.10\times}$ & \textbf{8.29} & $\mathbf{7.59\times}$ & \textbf{9.66} & $\mathbf{5.35\times}$ & \textbf{7.42} & $\mathbf{6.96\times}$ & \textbf{8.80} & $\mathbf{7.07\times}$ & \textbf{8.76} & $\mathbf{7.31\times}$ & \textbf{9.31} & $\mathbf{4.35\times}$ & \textbf{6.29} & $\mathbf{3.36\times}$ & \textbf{4.58} & $\mathbf{6.01\times}$ & \textbf{7.89} \\
\cmidrule{1-20}
& EAGLE-3 & $2.86\times$ & 2.52 & $2.76\times$ & 2.35 & $2.49\times$ & 2.12 & $2.63\times$ & 2.49 & $2.52\times$ & 2.23 & $2.28\times$ & 1.84 & $2.45\times$ & 2.04 & $2.02\times$ & 1.65 & $2.50\times$ & 2.16 \\
& DFlash & $4.60\times$ & 5.90 & $4.53\times$ & 6.70 & $3.64\times$ & 4.94 & $4.39\times$ & 5.64 & $4.15\times$ & 5.42 & $5.23\times$ & 7.09 & $2.47\times$ & 4.01 & $2.16\times$ & 2.84 & $3.90\times$ & 5.32 \\
\rowcolor{blue!10} \cellcolor{white} \multirow{-3}{*}{Q3-8B} & \textbf{\algo} & $\mathbf{5.51\times}$ & \textbf{8.37} & $\mathbf{6.71\times}$ & \textbf{9.12} & $\mathbf{5.29\times}$ & \textbf{6.96} & $\mathbf{6.29\times}$ & \textbf{8.32} & $\mathbf{5.94\times}$ & \textbf{8.20} & $\mathbf{6.38\times}$ & \textbf{9.40} & $\mathbf{3.69\times}$ & \textbf{6.02} & $\mathbf{3.30\times}$ & \textbf{4.55} & $\mathbf{5.39\times}$ & \textbf{7.62} \\
\bottomrule
\end{tabular}
}
\end{table*}

%% file: figures/addition_gpu_model_results.tex
\begin{figure}[t]
    \centering
    \begin{subfigure}[t]{\textwidth}
        \centering
        \includegraphics[width=0.4\textwidth]{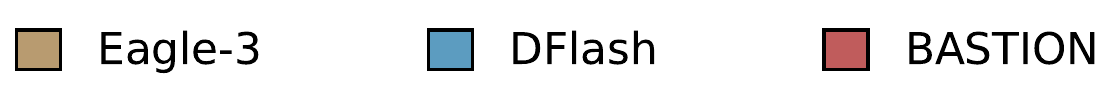}
    \end{subfigure}
    \centering
    \begin{subfigure}[t]{0.353\textwidth}
        \includegraphics[width=\textwidth]{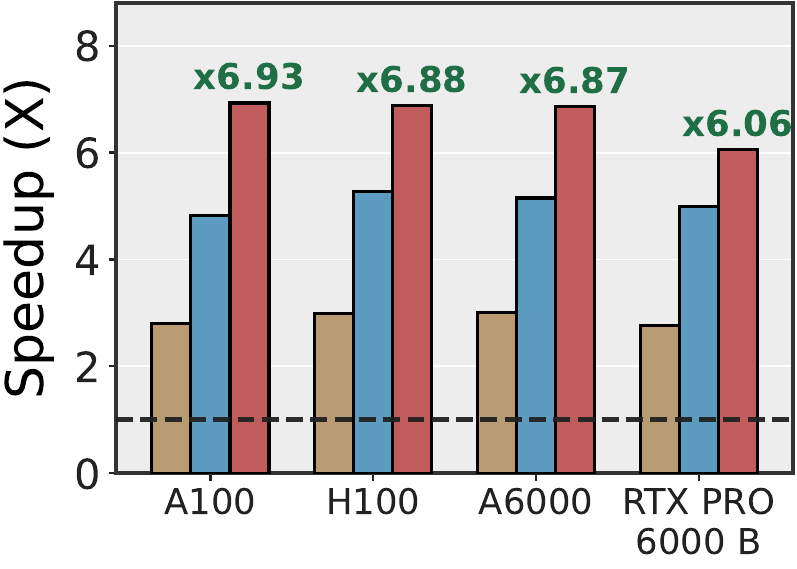}
        \vspace{-15pt}
        \subcaption{Qwen3-4B}
    \end{subfigure}
    \hfill
    \centering
    \begin{subfigure}[t]{0.306\textwidth}
        \includegraphics[width=\textwidth]{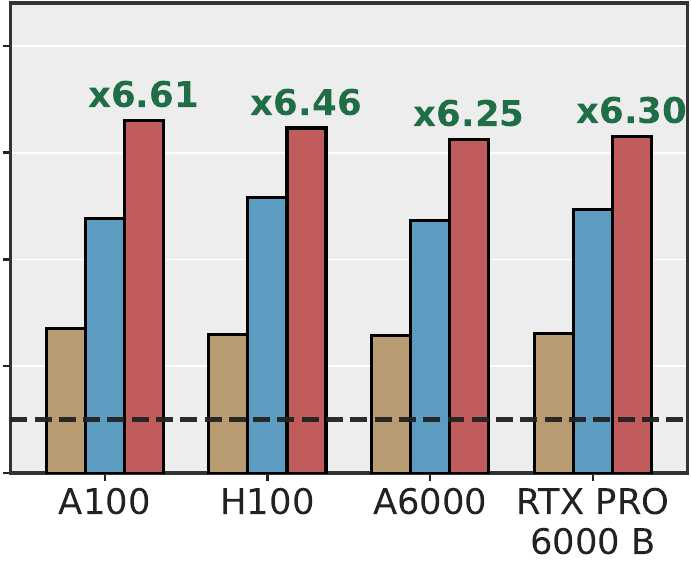}
        \vspace{-15pt}
        \subcaption{Qwen3-8B}
    \end{subfigure}
    \hfill
    \centering
    \begin{subfigure}[t]{0.306\textwidth}
        \includegraphics[width=\textwidth]{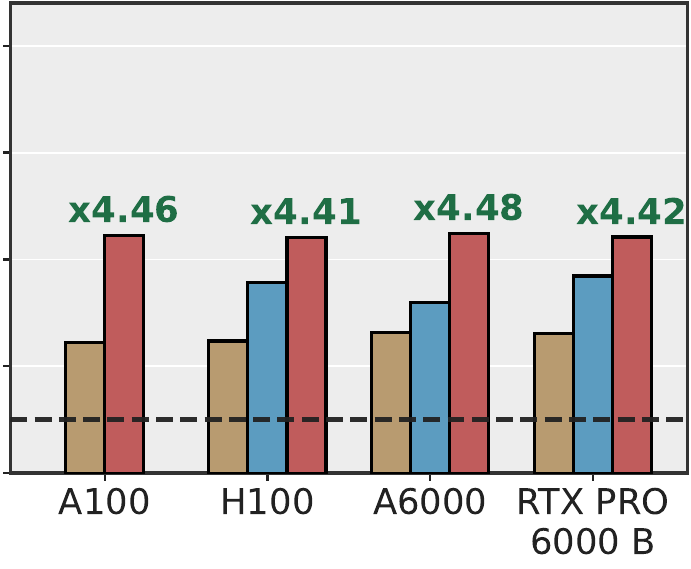}
        \vspace{-15pt}
        \subcaption{Llama-3.1-8B-Instruct}
    \end{subfigure}
    \caption{\textbf{Additional speedup results across GPU architectures.}
    Per-cell average wall-clock speedup of \algo{} versus EAGLE-3 and DFlash on
    (a) Qwen3-4B, (b) Qwen3-8B, and (c) Llama-3.1-8B-Instruct, evaluated on four
    NVIDIA GPUs (A100, H100, A6000, and RTX PRO 6000 Blackwell) at temperature
    $T=0$. Each bar reports the mean speedup over autoregressive decoding across
    all eight benchmarks. Numbers above each red bar give the speedup achieved
    by \algo{}, which dominates both baselines on every (model, GPU) cell.
    \looseness=-1
    }
    \label{fig:additional_results}
\end{figure}

%% file: tables/best_first_vs_beam_search.tex
\begin{figure*}[!t]
\vspace{-8pt}
    \centering
    % --- Left: (a) Figure ---
    \begin{minipage}[c]{0.48\textwidth}
        \centering
        \includegraphics[width=\linewidth]{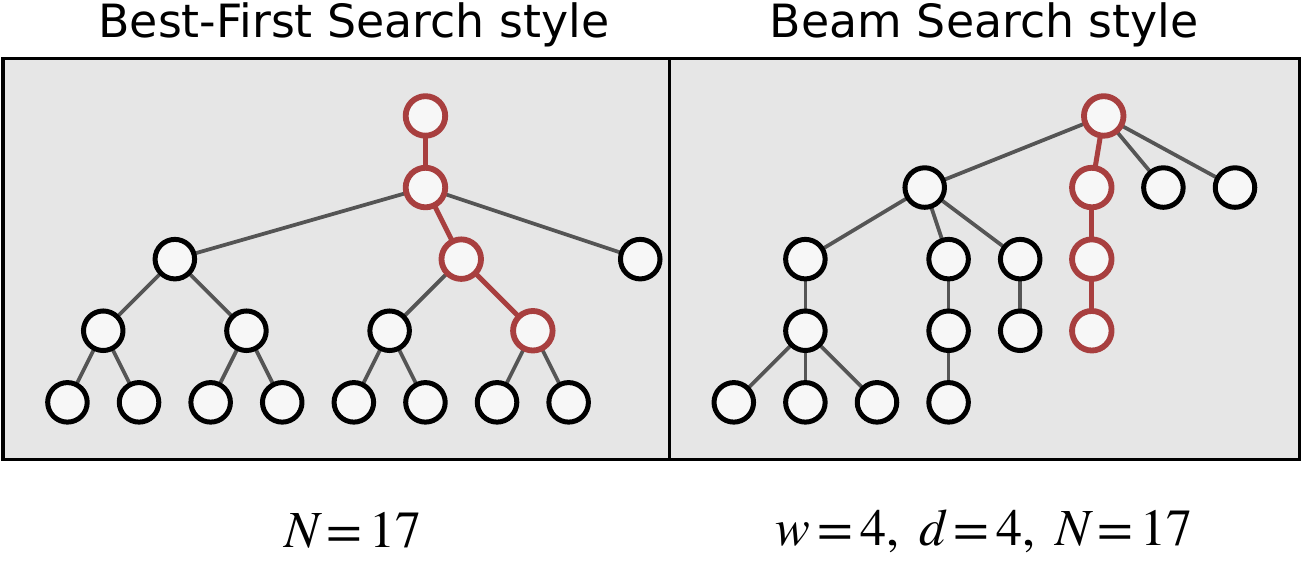}
    \subcaption{Tree topology under $N=17$}
    \end{minipage}\hfill
    % --- Right: (b) Table ---
    \begin{minipage}[c]{0.48\textwidth}
    \centering
    {\small
    \setlength{\tabcolsep}{3pt}
    \begin{tabular}{l | l c | c c} 
    \toprule
    \textbf{Model} & \textbf{Method} & \textbf{Tree} & \textbf{Speedup} & $\mathbf{\tau}$ \\
    \midrule
    \multirow{3}{*}{Q3-4B} & greedy~\citep{chen2026dflash} & \textcolor{red}{\xmark} & $5.14\times$ & 6.10 \\
     & beam search & \textcolor{green!60!black}{\checkmark} & $6.16\times$ & 7.77 \\
     & \cellcolor{blue!10}best-first & \cellcolor{blue!10}\textcolor{green!60!black}{\checkmark} & \cellcolor{blue!10}$\mathbf{6.59\times}$ & \cellcolor{blue!10}$\mathbf{8.19}$ \\
    \midrule
    \multirow{3}{*}{Q3-8B} & greedy~\citep{chen2026dflash} & \textcolor{red}{\xmark} & $4.73\times$ & 6.09 \\
     & beam search & \textcolor{green!60!black}{\checkmark} & $5.74\times$ & 7.60 \\
     & \cellcolor{blue!10}best-first & \cellcolor{blue!10}\textcolor{green!60!black}{\checkmark} & \cellcolor{blue!10}$\mathbf{6.09\times}$ & \cellcolor{blue!10}$\mathbf{7.96}$ \\
    \bottomrule
    \end{tabular}}
    \subcaption{Average speedup across 8 benchmarks}
\end{minipage}
    \caption{\textbf{Tree expansion under fixed budgets.}
    \textbf{(a)} At $N{=}17$, beam search spreads nodes uniformly, while best-first focuses on high-scoring prefixes (red: accepted prefix).
\textbf{(b)} Average A6000 speedup across 8 math/code/chat benchmarks. Under matched budgets, best-first ($N{=}61$) outperforms beam ($w{=}4,d{=}15$), improving Qwen3-4B/8B by $+7.0\%$/$+6.1\%$ (higher $\tau$). Greedy~\citep{chen2026dflash} (single-path, block $16$) is an unmatched no-tree baseline. \textbf{Bold} indicates the best per block.
    % \textbf{(a)} At $N{=}17$, beam search spreads nodes uniformly across depths while best-first concentrates them on high-scoring prefixes (red: target-accepted prefix).
    % \textbf{(b)} Average speedup across 8 math/code/chat benchmarks on A6000. Beam ($w{=}4,d{=}15$) and best-first ($N{=}61$) are budget-matched; best-first improves Qwen3-4B/8B speedup by $+7.0\%$/$+6.1\%$ with higher $\tau$. Greedy~\citep{chen2026dflash} (single-path, block size $16$) is a no-tree reference and \emph{not} budget-matched. \textbf{Bold}: best per block.
    }
    \label{fig:tree_style_comparison}
\end{figure*}

%% file: tables/sglang_throughput.tex
\begin{table}[!t]
\centering
\captionsetup{position=top,skip=7pt}
\caption{\textbf{Throughput under SGLang dynamic batching.} Qwen3-8B on an NVIDIA A6000 at $T=0$, averaged over HumanEval and MATH500; throughput is in tokens/s. $C$ is the number of in-flight requests rather than a fixed batch size: SGLang dynamically batches runnable requests, so $C$ bounds an effective batch size that varies during serving. A single calibration is reused across both tasks and all concurrency levels, with no refitting or budget tuning. \textbf{Bold} indicates the best value in each column; the highlighted row is \algo.}
\label{tab:sglang_throughput}
\small
\setlength{\tabcolsep}{5pt}
\renewcommand{\arraystretch}{1.05}
\begin{tabular}{@{}l cccccc cc@{}}
\toprule
\multirow{2}{*}{\textbf{Method}} & \multicolumn{6}{c}{\textbf{Throughput (tokens/s) at request concurrency $C$}} & \multicolumn{2}{c}{\textbf{Average}} \\
\cmidrule(lr){2-7} \cmidrule(lr){8-9}
 & $C{=}1$ & $C{=}4$ & $C{=}8$ & $C{=}16$ & $C{=}32$ & $C{=}64$ & tok/s & $\tau$ \\
\midrule
DFlash                    & 208 & 679 & 1{,}088 & 1{,}310 & 1{,}476 & 1{,}494 & 1{,}042 & 7.29 \\
\algo-Fixed ($N{=}16$)    & 227 & 714 & $\mathbf{1{,}148}$ & 1{,}381 & $\mathbf{1{,}593}$ & 1{,}578 & $\mathbf{1{,}107}$ & 8.00 \\
\algo-Fixed ($N{=}32$)    & $\mathbf{239}$ & $\mathbf{719}$ & 924 & 1{,}112 & 1{,}170 & 1{,}180 & 891 & 8.70 \\
\algo-Fixed ($N{=}64$)    & 233 & 514 & 623 & 661 & 697 & 708 & 573 & $\mathbf{9.22}$ \\
\rowcolor{blue!10} \textbf{\algo} & 237 & 712 & 1{,}096 & $\mathbf{1{,}389}$ & 1{,}520 & $\mathbf{1{,}597}$ & 1{,}092 & 7.65 \\
\bottomrule
\end{tabular}
\end{table}

%% file: tables/sglang_per_step_latency.tex
\begin{table}[!htbp]
\centering
\captionsetup{position=top,skip=7pt}
\caption{\textbf{Per-step latency decomposition at effective batch ${\approx}\,36$.} Same setup as Table~\ref{tab:sglang_throughput} (Qwen3-8B, A6000, $T=0$, SGLang dynamic batching). \emph{Preparation} covers drafting and tree construction; \emph{verification} is the target forward pass. Under batching, the fixed tree pays $+20.5\%$ step time for only $+8.9\%$ accepted tokens.}
\label{tab:sglang_per_step}
\small
\setlength{\tabcolsep}{6pt}
\renewcommand{\arraystretch}{1.05}
\begin{tabular}{@{}l cccc@{}}
\toprule
\textbf{Method} & \textbf{Preparation} & \textbf{Verification} & \textbf{Total step} & $\bm{\tau}$ \\
\midrule
DFlash (chain, block $16$) & $16.94$\,ms & $95.90$\,ms & $115.84$\,ms & $7.71$ \\
\algo-Fixed ($N{=}16$)     & $25.29$\,ms & $110.33$\,ms & $139.57$\,ms & $8.40$ \\
\midrule
\textbf{Increase}          & $\mathbf{+49.3\%}$ & $\mathbf{+15.1\%}$ & $\mathbf{+20.5\%}$ & $\mathbf{+8.9\%}$ \\
\bottomrule
\end{tabular}
\end{table}

%% file: tex/06_analysis.tex
\section{Analysis}

% \subsection{Fixed vs. Adaptive Budget Policies}
% \label{subsec:fixed_budget_sweep}

% \input{figures/budget_speedup}

% \paragraph{\algo-Fixed exposes strong budget sensitivity.}
% We first sweep the verification budget $N$ for \algo-Fixed (best-first expansion with manually chosen $N$). \autoref{fig:fixed_budget_compare} shows that the resulting speedup curves are sharply peaked: small $N$ underutilizes the verifier, large $N$ inflates verification latency faster than it improves accepted length, and the peak location varies across models, GPUs, and benchmarks. A fixed budget is therefore a workload- and hardware-dependent system decision.

% \vspace{-6pt}
% \paragraph{\algo-Adaptive selects near-efficient budgets without per-setting tuning.}
% \algo-Adaptive replaces the manual choice with an online estimate of the acceptance--latency trade-off. Across all panels in \autoref{fig:fixed_budget_compare}, the adaptive variant tracks the per-setting Oracle---the upper bound any fixed $N$ could achieve in hindsight---eliminating the need for a per-setting sweep by construction.

\subsection{Fixed vs. Adaptive Budget Policies}
\label{subsec:fixed_budget_sweep}

\input{figures/budget_speedup}

\paragraph{Fixed-budget tree construction exposes strong budget sensitivity.}
We first sweep the verification budget $N$ for the fixed-budget variant (best-first expansion with manually chosen $N$; denoted \emph{Fixed} in \autoref{fig:fixed_budget_compare}). The resulting speedup curves are sharply peaked: small $N$ underutilizes the verifier, large $N$ inflates verification latency faster than it improves accepted length, and the peak location varies across models, GPUs, and benchmarks. A fixed budget is therefore a workload- and hardware-dependent system decision.

\vspace{-6pt}
\paragraph{\algo selects near-efficient budgets without per-setting tuning.}
\algo replaces the manual choice with an online estimate of the acceptance--latency trade-off. Across all panels in \autoref{fig:fixed_budget_compare}, \algo tracks the per-setting Oracle---the upper bound any fixed $N$ could achieve in hindsight---eliminating the need for a per-setting sweep by construction.

\subsection{Latency Model Evaluation}
\label{subsec:latency_eval}

\input{figures/online_calibration_analysis}

\vspace{-6pt}
\paragraph{Calibrated roofline as the default latency estimator.}
\autoref{fig:latency_calib} compares measured and predicted verification latency for Qwen3-8B and Llama-3.1-8B at context lengths $c\in\{64,256,1024\}$. The bare roofline captures the scaling trend but underestimates absolute latency; step-function surges reflect CUDA tile-allocation overhead. Per-(GPU, target) linear calibration reduces RMSE by $87$--$90\%$ ($30.6{\to}3.1$\,ms and $26.4{\to}3.5$\,ms, respectively). This calibrated estimator is the default in \algo, denoted \emph{Static} below.

\vspace{-6pt}
\paragraph{Online-calibration ablation.}
We test whether runtime feedback can replace or refine offline calibration. \emph{EMA} tracks a multiplicative bias $\gamma$ on the bare roofline, $\gamma\leftarrow(1{-}\alpha)\gamma+\alpha(\text{obs}/\text{pred})$, and predicts latency as $\gamma R_t(N)$. \emph{EMA+Calib} instead applies this EMA as a residual correction to the offline-calibrated curve.

\vspace{-6pt}
\paragraph{Static is the most robust default; EMA alone is fragile.}
\autoref{fig:online_variants} compares mean speedup, acceptance length $\tau$, and tree size $\bar{N}$ across eight short-context benchmarks on A100 at $T{=}0$. \emph{Static} is best on Llama-3.1-8B and within $0.28\times$ of the best on Qwen3. \emph{EMA} gains $+0.28\times$/$+0.18\times$ with tighter Qwen3 trees, but larger Llama-3.1-8B trees yield little acceptance gain and reduce speedup by $12.7\%$. \emph{EMA+Calib} stays within $\pm0.06\times$ of \emph{Static}, indicating limited residual-correction benefit. We default to \emph{Static}; \emph{EMA} works without offline calibration but has model-dependent performance, whereas \emph{EMA+Calib} requires the calibrated curve.

Appendix~\ref{app:subsec:calib_cost} reports the profiling and fitting costs and discusses when recalibration is needed.

%% file: figures/budget_speedup.tex
\begin{figure}[!t]
    \centering
    \small
    \includegraphics[width=0.5\textwidth]{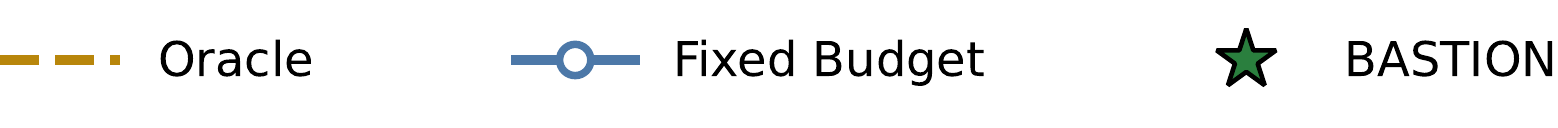} 
    \vspace{-3pt}
    \includegraphics[width=\textwidth]{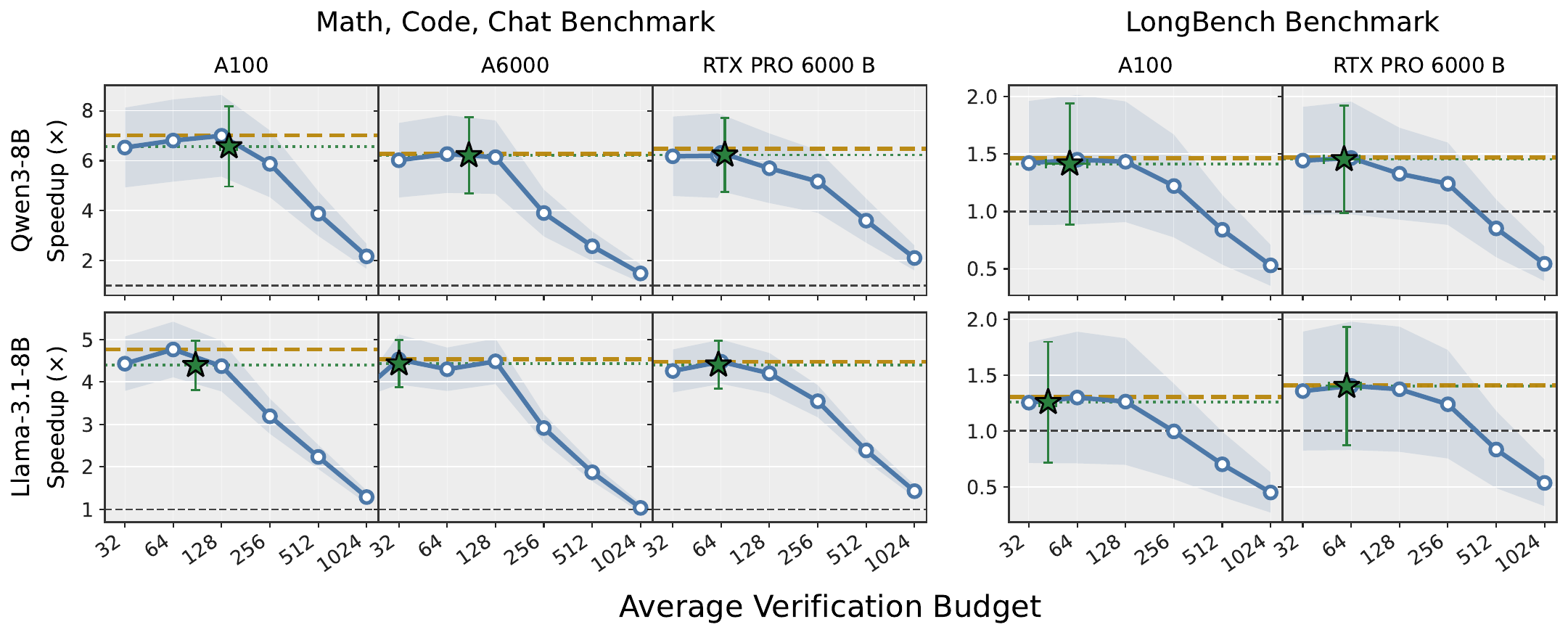}
    \vspace{-15pt}    
    \caption{
    % \textbf{Budget-policy sweep within \algo.} Mean speedup over AR decoding at $T{=}0$. \textbf{Blue:} \algo-Fixed with $N{\in}\{32,64,128,256,512,1024\}$. \textbf{Green stars:} \algo-Adaptive at its mean realized budget. \textbf{Dashed gold (\emph{Oracle}):} per-dataset best fixed $N$ averaged within each panel---an upper bound on any static $N$ without per-dataset tuning. \emph{Left:} short-context benchmarks over $\{\text{A100,B6000,A6000}\}{\times}\{\text{Qwen3-8B, Llama-3.1-8B-Instruct}\}$; \emph{Right:} LongBench (English) over $\{\text{B6000,A100}\}{\times}$ same targets. \algo tracks the Oracle on most panels, recovering near-best fixed-budget speedup without tuning, while over-budgeting sharply degrades the fixed sweep.
    \textbf{Budget-policy sweep within \algo.} Mean speedup over AR decoding at $T{=}0$. \textbf{Blue:} \algo-Fixed ($N{\in}\{32,64,128,256,512,1024\}$). \textbf{Green stars:} \algo (mean realized budget). \textbf{Dashed gold (\emph{Oracle}):} best per-dataset fixed $N$ averaged per panel---an upper bound for static $N$ without tuning. \emph{Left:} short-context benchmarks over $\{\text{A100,A6000,RTX PRO 6000 B}\}{\times}\{\text{Qwen3-8B, Llama-3.1-8B-Instruct}\}$; \emph{Right:} LongBench (English) over $\{\text{A100,RTX PRO 6000 B}\}{\times}$ same targets. \algo tracks the Oracle on most panels, recovering near-best speedup without tuning, while over-budgeting degrades the fixed sweep.
    }
    \label{fig:fixed_budget_compare}
\end{figure}

%% file: figures/online_calibration_analysis.tex
\begin{figure*}[!t]
    \centering
    % --- Row 1: images (left) and table (right), vertically centered ---
    \begin{minipage}[c]{0.54\textwidth}
        \centering
        % 왼쪽 상단: Legend
        \includegraphics[width=0.55\linewidth]{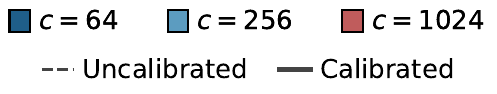}
        
        % 왼쪽 하단: calibration_a 와 calibration_b 나란히 배치
        \begin{minipage}[c]{0.54\linewidth}
            \includegraphics[width=\linewidth]{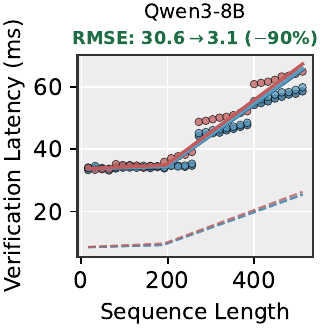}
        \end{minipage}\hfill
        \begin{minipage}[c]{0.43\linewidth}
            \includegraphics[width=\linewidth]{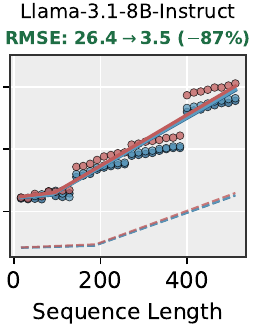}
        \end{minipage}
    \end{minipage}\hfill
    \begin{minipage}[c]{0.45\textwidth}
        \centering
        \small
        \setlength{\tabcolsep}{2pt}
        \begin{tabular}{l | l | c c c}
        \toprule
        \textbf{Target} & \textbf{Method} & \textbf{Speedup} & $\tau$ & $\bar{N}$ \\
        \midrule
        \multirow{3}{*}{Q3-4B}
         & \cellcolor{blue!10}Static
           & \cellcolor{blue!10}$\underline{6.93\times}$
           & \cellcolor{blue!10}$\underline{8.55}$
           & \cellcolor{blue!10}$202.0$ \\
         & EMA+Calib & $6.87\times$ & $\mathbf{8.56}$ & $201.9$ \\
         & EMA        & $\mathbf{7.21\times}$ & $8.22$ & $102.2$ \\
        \midrule
        \multirow{3}{*}{Q3-8B}
         & \cellcolor{blue!10}Static
           & \cellcolor{blue!10}$\underline{6.61\times}$
           & \cellcolor{blue!10}$\mathbf{8.47}$
           & \cellcolor{blue!10}$139.3$ \\
         & EMA+Calib & $6.55\times$ & $8.44$ & $139.8$ \\
         & EMA        & $\mathbf{6.79\times}$ & $\underline{8.45}$ & $126.6$ \\
        \midrule
        \multirow{3}{*}{L3.1-8B}
         & \cellcolor{blue!10}Static
           & \cellcolor{blue!10}$\mathbf{4.46\times}$
           & \cellcolor{blue!10}$\underline{6.00}$
           & \cellcolor{blue!10}$87.0$ \\
         & EMA+Calib & $\underline{4.42\times}$ & $5.99$ & $87.0$ \\
         & EMA        & $3.89\times$ & $\mathbf{6.13}$ & $123.1$ \\
        \bottomrule
        \end{tabular}
    \end{minipage}

    \vspace{0.5em}

    % --- Row 2: sub-captions, top-aligned, with referenceable labels ---
    \begin{minipage}[t]{0.54\textwidth}
        \centering
        \subcaption{Verification-latency calibration}\label{fig:latency_calib}
    \end{minipage}\hfill
    \begin{minipage}[t]{0.45\textwidth}
        \centering
        \subcaption{Online-calibration variants}\label{fig:online_variants}
    \end{minipage}

    \caption{\textbf{Latency model evaluation on A100.}
    \textbf{(a)} Verification latency vs. sequence length for two targets at contexts $c\in\{64,256,1024\}$. Dashed and solid lines denote the uncalibrated roofline and calibrated fit (used by the controller). Calibration cuts RMSE by $87$--$90\%$.
    \textbf{(b)} Mean over 8 short-context benchmarks at $T{=}0$ ($\bar{N}$: mean realized tree size). \algo variants: \emph{Static} (offline curve), \emph{EMA+Calib} (offline + online EMA residual), and \emph{EMA} (online only). \textbf{Bold} and \underline{underline} indicate the best and second best per model. Highlighted: default \algo config.}
    \label{fig:latency_eval}
\end{figure*}

%% file: tex/06_conclusion.tex
\section{Conclusion}
We presented \algo, a budget-aware tree-based speculative decoding method for block-diffusion drafters. \algo combines a path-confidence surrogate with provably optimal best-first tree construction (Proposition~\ref{prop:bestfirst}) and a hardware-calibrated online budget controller (Proposition~\ref{prop:unimodal}). Across three targets, four GPUs, and diverse benchmarks, \algo achieves up to $6.61\times$ speedup over autoregressive decoding ($1.39\times$ over DFlash, $2.45\times$ over EAGLE-3) without per-deployment tuning.

%% file: tex/07_appendix.tex
%======================================================================
% APPENDIX
%======================================================================
\newpage
\appendix
\onecolumn

\setlength{\cftbeforesecskip}{16pt}
\setlength{\cftbeforesubsecskip}{4pt}
\renewcommand\cftsecpagefont{\color{Crimson}}
\renewcommand\cftsubsecpagefont{\color{Crimson}}
\renewcommand\cftsubsubsecpagefont{\color{Crimson}}
% \renewcommand{\contentsname}{\large{Contents}}
% {
%   \hypersetup{linkcolor=}
%   \tableofcontents
% }

\clearpage
\input{tex/appendix/07_limitations}

\input{tex/02_related_works}

\clearpage
\input{tex/appendix/01_math_proofs_new}
\clearpage
\input{tex/appendix/02_algorithm}
\clearpage
\input{tex/appendix/03_overal_pipeline}
\clearpage
\input{tex/appendix/04_additional_exps}
\clearpage
\input{tex/appendix/05_cost_models}

\clearpage
\input{tex/appendix/06_experimental_setup}

%% file: tex/appendix/07_limitations.tex
\section{Limitations}
\label{app:sec:limits}

There are two limitations in our work:
\begin{itemize}[leftmargin=*, itemsep=2pt]
    \item \textbf{Batch size constraints:} Most evaluations target batch size~1. Section~\ref{subsec:sglang} presents an initial study under SGLang dynamic batching, but a broader treatment of multi-request serving remains future work.
    \item \textbf{Runtime profile calibration:} Our default offline calibration assumes a stable runtime profile; while \algo with EMA from static calibration provides a residual-correction fallback, deployments with pronounced runtime drift remain an open direction.
\end{itemize}
The broader framework---surrogate-driven tree construction with online budget control---extends naturally to other multi-token prediction paradigms beyond block diffusion.

%% file: tex/02_related_works.tex
\section{Related Works}
\label{sec:related_work}

% \paragraph{Block-diffusion drafters for speculative decoding.}
% Block-diffusion and diffusion-inspired drafters reduce draft latency by predicting multiple future token distributions in parallel rather than rolling out tokens sequentially. DFlash~\citep{chen2026dflash} uses a block-diffusion drafter for speculative decoding, while DART~\citep{liu2026dart} and TiDAR~\citep{liu2025tidar} explore related diffusion--autoregressive hybrid decoding strategies. These methods show the potential of parallel draft generation, but they leave open how to allocate the target-model verification budget when many plausible continuations are available.

\paragraph{Block diffusion drafter for speculative decoding.}
Recent speculative decoding methods have transitioned from autoregressive drafting~\citep{chen2023accelerating, leviathan2023fast, kim2023speculative, li2024eagle} toward multi-token prediction and block diffusion paradigms~\citep{cai2024medusa, samragh2025your, li2025diffuspec, liu2025tidar, chen2026dflash}. By leveraging the inherent multi-token predictive capacity of modern LLMs~\citep{samragh2025your}, parallel drafting maximizes hardware utilization and reduces latency via single forward-pass generation. For example, PARD~\citep{an2025pard} converts autoregressive drafters into target-independent parallel models via a conditional drop-token mechanism, while DiffuSpec~\citep{li2025diffuspec} enables training-free integration of pretrained diffusion models using causal-consistency path search. Other works directly pretrain block diffusion drafters: TiDAR~\citep{liu2025tidar} synergizes diffusion-style drafting with causal verification, and DFlash~\citep{chen2026dflash} leverages the target model's context features~\citep{li2025eagle} via KV injection.
\looseness=-1

% \vspace{-6pt}
% \paragraph{Tree-based speculative decoding.}

% Tree-based speculative decoding improves verifier utilization by replacing a single draft trajectory with a prefix tree whose root-to-leaf paths represent alternative continuations. SpecInfer~\citep{miao2024specinfer} verifies such trees through tree attention, while Medusa~\citep{cai2024medusa}, EAGLE~\citep{li2024eagle}, and EAGLE-2~\citep{li2024eagle2fasterinferencelanguage} construct multi-token or tree-structured drafts using auxiliary heads or feature-level draft models. Concurrent with our work, DDTree~\citep{ringel2026accelerating} independently studies fixed-budget tree construction for block-diffusion drafters using drafter path scores and best-first expansion. These works demonstrate that tree structure is critical for acceptance efficiency, but they typically fix the tree size or choose it through external tuning. This limits portability: a budget tuned for one GPU, model, or dataset can become suboptimal when verifier latency or acceptance statistics change.

\vspace{-6pt}
\paragraph{Tree-based speculative decoding.}

Tree-based speculative decoding improves verifier utilization by replacing a single draft trajectory with a prefix tree, where root-to-leaf paths represent alternative continuations. SpecInfer~\citep{miao2024specinfer} pioneers the verification of such trees through tree attention, while methods like Medusa~\citep{cai2024medusa}, EAGLE~\citep{li2024eagle}, and EAGLE-2~\citep{li2024eagle2fasterinferencelanguage} construct multi-token or tree-structured drafts using auxiliary heads or feature-level autoregressive models. To maximize expected acceptance length under hardware constraints, approaches like Sequoia~\citep{chen_sequoia_2025} and SpecTr~\citep{sun2023spectr} optimize fixed tree topologies offline through exact search or dynamic programming algorithms using profiling statistics. This static-tree paradigm has also been extended to hybrid state-space models~\citep{wu2025stree}.
\looseness=-1

\vspace{-6pt}
\paragraph{Adaptive draft tree structure.}

Recent methods further optimize acceptance efficiency by dynamically adjusting draft tree topologies~\citep{huang2024specdec++, liu2024pearl, wang2025opt, guan2025yggdrasilbridgingdynamicspeculation, liu2025logitspec}. While Sequoia~\citep{chen_sequoia_2025} relies on offline profiling, subsequent approaches introduce online adaptation by modifying draft structures via expectation maximization (OPT-Tree~\citep{wang2025opt}), Markov decision processes (SpecDec++~\citep{huang2024specdec++}), parallel serving (PEARL~\citep{liu2024pearl}), length predictors (AdaEAGLE~\citep{zhang2024adaeagle}), or latency-aware optimization (Yggdrasil~\citep{guan2025yggdrasilbridgingdynamicspeculation}). 
Concurrent with our work, DDTree~\citep{ringel2026accelerating} also extends this adaptability to block diffusion drafters via best-first tree expansion. Our work shares a similar tree-based block diffusion drafting paradigm; however, whereas DDTree relies on fixed budgets and manual tuning, we further introduce a hardware-aware adaptive controller to dynamically align the draft structure with real-time system constraints. Appendix~\ref{app:subsec:ddtree} presents direct experimental comparisons with DDTree under matched fixed budgets and evaluates the benefit of our adaptive budget control.
\looseness=-1

%% file: tex/appendix/01_math_proofs_new.tex
% ==========================================================================
% Appendix B — Mathematical Proofs
% ==========================================================================
\section{Mathematical Proofs}
\label{app:proofs}
 
% --------------------------------------------------------------------------
\subsection{Proofs of Lemma~\ref{lem:cover-count} and
            Proposition~\ref{prop:path-sum}}
\label{app:proofs:b1}
 
We adopt the notation of \S\ref{sec:method:surrogate}. Let $\mathcal{T}$ be a
prefix-closed tree with root $r$, and let $X = (X_1,\dots,X_\gamma)$ denote a
sample drawn from the drafter's joint distribution; by property (P2),
$X_k \sim q_k(\cdot)$ independently across positions. For a node
$i\in\mathcal{T}$ at depth $d(i)$ with root-to-node path
$\mathrm{path}(i) = (x_{i_1},\dots,x_{i_{d(i)}})$, define the \emph{cover
indicator}
\[
\mathrm{cov}_i(X)
\;:=\;
\mathbf{1}\!\left\{X_{1:d(i)} = \mathrm{path}(i)\right\},
\]
with the convention $\mathrm{cov}_r(X) = 1$ (the empty prefix is vacuously
matched). Let $\mathcal{C}(X) := \{i\in\mathcal{T} : \mathrm{cov}_i(X)=1\}$
denote the covered set. Under the self-verification model, a node is
committed iff it is covered, so the committed length is
$A_{\mathrm{self}}(\mathcal{T};X) = |\mathcal{C}(X)|$.
 
\subsubsection*{Proof of Lemma~\ref{lem:cover-count}}
 
We establish three structural properties of $\mathcal{C}(X)$, from which the
lemma follows.
 
\textbf{(i) Ancestor-closure.}
Let $i\in\mathcal{C}(X)$ and let $j$ be an ancestor of $i$ in $\mathcal{T}$.
Since $j$ lies on the root-to-$i$ path, $\mathrm{path}(j)$ is a prefix of
$\mathrm{path}(i)$, so $\mathrm{path}(i)_{1:d(j)} = \mathrm{path}(j)$.
Therefore
\[
X_{1:d(j)}
\;=\; \mathrm{path}(i)_{1:d(j)}
\;=\; \mathrm{path}(j),
\]
which gives $j\in\mathcal{C}(X)$.
 
\textbf{(ii) At most one covered node per depth.}
Suppose two distinct nodes $i_1,i_2\in\mathcal{T}$ at depth $k$ both lie in
$\mathcal{C}(X)$. Then $X_{1:k} = \mathrm{path}(i_1)$ and
$X_{1:k} = \mathrm{path}(i_2)$, forcing
$\mathrm{path}(i_1) = \mathrm{path}(i_2)$. But in a tree, two distinct nodes
at the same depth necessarily have distinct root-to-node paths: tracing from
the root, their paths must agree up to their nearest common ancestor and
then diverge into different children. Contradiction.
 
\textbf{(iii) Chain structure.}
Set $m := \max_{i\in\mathcal{C}(X)} d(i)$; the maximum is well-defined
because $r\in\mathcal{C}(X)$. By (ii), there exists a unique node
$i^\star\in\mathcal{C}(X)$ with $d(i^\star)=m$. By (i), every ancestor of
$i^\star$ in $\mathcal{T}$ also belongs to $\mathcal{C}(X)$, contributing
exactly one covered node at each depth $k\in\{0,1,\dots,m\}$. By (ii), no
node outside the root-to-$i^\star$ chain can be in $\mathcal{C}(X)$. Hence
$\mathcal{C}(X)$ is exactly the chain from $r$ to $i^\star$, and
$|\mathcal{C}(X)| = m+1$.
 
Combining the three properties,
\[
A_{\mathrm{self}}(\mathcal{T};X)
\;=\; |\mathcal{C}(X)|
\;=\; \#\!\left\{i\in\mathcal{T} : X_{1:d(i)} = \mathrm{path}(i)\right\}.
\qed
\]
 
\subsubsection*{Proof of Proposition~\ref{prop:path-sum}}
 
By Lemma~\ref{lem:cover-count}, the committed length decomposes as a sum of
cover indicators:
\[
A_{\mathrm{self}}(\mathcal{T};X)
\;=\; \sum_{i\in\mathcal{T}} \mathrm{cov}_i(X).
\]
Taking expectations over $X$ and applying linearity,
\[
\mathbb{E}_X\!\left[A_{\mathrm{self}}(\mathcal{T};X)\right]
\;=\; \sum_{i\in\mathcal{T}} \mathbb{E}_X[\mathrm{cov}_i(X)]
\;=\; \sum_{i\in\mathcal{T}}
      \mathbb{P}\!\left(X_{1:d(i)} = \mathrm{path}(i)\right).
\]
For the root, $\mathbb{P}(X_{1:0}=\mathrm{path}(r)) = 1 = \rho(r)$ by
convention. For any non-root $i$ with
$\mathrm{path}(i) = (x_{i_1},\dots,x_{i_{d(i)}})$, position-wise independence
(P2) yields
\[
\mathbb{P}\!\left(X_{1:d(i)} = \mathrm{path}(i)\right)
\;=\; \prod_{k=1}^{d(i)} \mathbb{P}(X_k = x_{i_k})
\;=\; \prod_{k=1}^{d(i)} q_k(x_{i_k})
\;=\; \rho(i).
\]
Summing over $\mathcal{T}$,
\[
\widehat{A}(\mathcal{T})
\;:=\; \mathbb{E}_X\!\left[A_{\mathrm{self}}(\mathcal{T};X)\right]
\;=\; \sum_{i\in\mathcal{T}} \rho(i),
\]
which establishes \Eqref{eq:surrogate}. \qed

% --------------------------------------------------------------------------
\subsection{Proof of Proposition~\ref{prop:bestfirst}}
\label{app:proofs:b2}
 
Let $i_1, i_2, \dots$ be the nodes added by best-first expansion in order,
with $i_1 = r$.
 
\textbf{Step 1: Path monotonicity.}
For any non-root $i$ with parent $\pi(i)$,
\[
\rho(i) \;=\; \rho(\pi(i))\cdot q_{d(i)}(x_i) \;\le\; \rho(\pi(i)),
\]
since $q_{d(i)}(x_i)\in[0,1]$.
 
\textbf{Step 2: Best-first selects in globally non-increasing order of
$\rho$.}
Suppose, for contradiction, that there exist indices $m<n$ with
$\rho(i_m) < \rho(i_n)$. Among such pairs, choose the one minimizing $m$. At
step $m$, best-first selected $i_m$ from the available frontier $F_m$ (nodes
whose parent has been added but which are not yet in $\mathcal{T}_{m-1}$).
Consider the root-to-$i_n$ path in the lattice $\mathcal{V}$, and let $w$ be
the \emph{first} node on this path that is not in $\{i_1,\dots,i_{m-1}\}$.
Then $\pi(w)\in\{i_1,\dots,i_{m-1}\}$, so $w\in F_m$. By Step~1 applied along
the path from $w$ down to $i_n$,
\[
\rho(w) \;\ge\; \rho(i_n) \;>\; \rho(i_m).
\]
But best-first chose $i_m$ as the available node of maximal $\rho$,
contradicting $\rho(w)>\rho(i_m)$ with $w\in F_m$. Hence
$\rho(i_1)\ge\rho(i_2)\ge\cdots$.
 
\textbf{Step 3: Optimality via leaf exchange.}
Intuitively, Step~2 already implies that $\mathcal{T}_N$ collects the
$N$ largest path scores reachable under prefix-closure; we now make this
rigorous by showing that any prefix-closed tree of size $N$ can be
transformed into $\mathcal{T}_N$ via a sequence of score-non-decreasing
leaf-for-leaf exchanges.

Define $\mathcal{T}_N := \{i_1,\dots,i_N\}$. Note $\mathcal{T}_N$ is
prefix-closed: for any $i_k$ with $k\le N$, best-first added $\pi(i_k)$
before $i_k$, so $\pi(i_k)\in\mathcal{T}_N$.
 
Let $\mathcal{T}\subseteq\mathcal{V}$ be any prefix-closed tree with
$|\mathcal{T}| = N$ and $\mathcal{T}\neq\mathcal{T}_N$. We construct a
prefix-closed $\mathcal{T}'$ with $|\mathcal{T}'|=N$,
$\widehat{A}(\mathcal{T}')\ge\widehat{A}(\mathcal{T})$, and
$|\mathcal{T}'\,\triangle\,\mathcal{T}_N| < |\mathcal{T}\,\triangle\,\mathcal{T}_N|$.
 
\emph{Selecting the exchange pair.}
Let $y\in\mathcal{T}\setminus\mathcal{T}_N$ be a node of maximum depth in
$\mathcal{T}\setminus\mathcal{T}_N$. Then $y$ has no children in
$\mathcal{T}$: any child $c$ of $y$ in $\mathcal{T}$ has $d(c)>d(y)$, so by
maximality $c\notin\mathcal{T}\setminus\mathcal{T}_N$, i.e.,
$c\in\mathcal{T}_N$; but prefix-closure of $\mathcal{T}_N$ then forces
$y=\pi(c)\in\mathcal{T}_N$, contradicting $y\in\mathcal{T}\setminus\mathcal{T}_N$.
Hence $y$ is a leaf of $\mathcal{T}$ and $\mathcal{T}\setminus\{y\}$ is
prefix-closed.
 
Let $s$ be the smallest index such that $i_s\notin\mathcal{T}$. Such $s$
exists because $|\mathcal{T}|=|\mathcal{T}_N|=N$ and
$\mathcal{T}\neq\mathcal{T}_N$. By minimality,
$\{i_1,\dots,i_{s-1}\}\subseteq\mathcal{T}$, and
$\pi(i_s)\in\{i_1,\dots,i_{s-1}\}\subseteq\mathcal{T}$.
 
\emph{Constructing $\mathcal{T}'$.}
Define $\mathcal{T}' := (\mathcal{T}\setminus\{y\})\cup\{i_s\}$. We verify
the required properties:
 
\emph{Prefix-closure.}
(a) Removing $y$ preserves prefix-closure since $y$ is a leaf.
(b) Adding $i_s$ requires $\pi(i_s)\in\mathcal{T}'$. We have
$\pi(i_s)\in\{i_1,\dots,i_{s-1}\}\subseteq\mathcal{T}\setminus\{y\}$ (using
$\pi(i_s)\in\mathcal{T}_N$ but $y\notin\mathcal{T}_N$, hence $\pi(i_s)\neq y$).
So $\pi(i_s)\in\mathcal{T}'$.
 
\emph{Score improvement.}
Since $i_s\in\mathcal{T}_N$ and $y\notin\mathcal{T}_N$, in best-first order
the index of $i_s$ is at most $N$ while the index of $y$ exceeds $N$. By
Step~2, $\rho(i_s)\ge\rho(y)$. Therefore
\[
\widehat{A}(\mathcal{T}')
\;=\; \widehat{A}(\mathcal{T}) - \rho(y) + \rho(i_s)
\;\ge\; \widehat{A}(\mathcal{T}).
\]
 
\emph{Symmetric difference decreases by 2.}
$\mathcal{T}'$ removes $y$ (which was in
$\mathcal{T}\setminus\mathcal{T}_N$) and adds $i_s$ (which was in
$\mathcal{T}_N\setminus\mathcal{T}$).
 
Iterating this exchange yields a finite sequence of prefix-closed trees of
size $N$ with non-decreasing $\widehat{A}$, terminating at $\mathcal{T}_N$.
Therefore $\widehat{A}(\mathcal{T}_N)\ge\widehat{A}(\mathcal{T})$.
 
\textbf{Step 4: Diminishing returns.}
\Eqref{eq:concave-A} is immediate from Step~2:
$\Delta\widehat{A}(N) = \rho(i_{N+1})$, and the global non-increasing order
gives $\rho(i_{N+2})\le\rho(i_{N+1})$. \qed

% --------------------------------------------------------------------------
\subsection{Proof of Proposition~\ref{prop:unimodal}}
\label{app:proofs:b3}
 
Let $S(N) = L_{\mathrm{AR}}\cdot\widehat{A}(N)/C(N)$ on
$\{1,\dots,N_{\max}\}$. Recall the assumptions:
\begin{itemize}
\item[(A1)] $\widehat{A}(N)$ is non-decreasing and concave
  (Proposition~\ref{prop:bestfirst}, \Eqref{eq:concave-A}).
\item[(A2)] $C(N) > 0$ is convex and non-decreasing.
\end{itemize}
Without loss of generality drop the positive constant $L_{\mathrm{AR}}$ and
study $f(N) = \widehat{A}(N)/C(N)$.
 
\textbf{Step 1: Continuous extension.}
Linearly interpolate $\widehat{A}$ and $C$ to functions on $[1,N_{\max}]$.
The interpolated $\widehat{A}$ remains concave and non-negative; the
interpolated $C$ remains convex and strictly positive. Both extensions are
continuous.
 
\textbf{Step 2: Quasiconcavity of $f$.}
A function is quasiconcave iff every upper level set is convex. For any
$c\ge 0$,
\[
f(x)\ge c
\;\iff\;
\widehat{A}(x) - c\,C(x) \;\ge\; 0.
\]
The function $g_c(x) := \widehat{A}(x) - c\,C(x)$ is concave: $\widehat{A}$
is concave by (A1), $-c\,C$ is concave because $c\ge 0$ and $C$ is convex by
(A2), and the sum of concave functions is concave. The upper level set
$\{x : g_c(x)\ge 0\}$ of a concave function is convex. Hence
$\{x : f(x)\ge c\}$ is convex for every $c$, so $f$ is quasiconcave on
$[1,N_{\max}]$.
 
\textbf{Step 3: Continuous unimodality.}
A continuous quasiconcave function on a real interval is unimodal: there
exists $x^\star\in[1,N_{\max}]$ such that $f$ is non-decreasing on
$[1,x^\star]$ and non-increasing on $[x^\star,N_{\max}]$. Otherwise, one
could find $x_1<x_2<x_3$ with $f(x_2)<\min(f(x_1),f(x_3))$, and the upper
level set at level $\min(f(x_1),f(x_3))$ would contain $x_1$ and $x_3$ but
not $x_2$, contradicting convexity.
 
\textbf{Step 4: Discrete unimodality.}
Restrict to integers. Let
$N^\star\in\arg\max_{N\in\{1,\dots,N_{\max}\}} S(N)$. By Step~3, $S$ is
non-decreasing on integers in $[1,x^\star]$ and non-increasing on integers
in $[x^\star,N_{\max}]$, so $S$ is non-decreasing on $\{1,\dots,N^\star\}$
and non-increasing on $\{N^\star,\dots,N_{\max}\}$.
 
\textbf{Step 5: First-decrease optimality.}
Suppose $N$ is the smallest index with $S(N{+}1) < S(N)$. Then $S$ has not
yet decreased on $\{1,\dots,N\}$, so $S(1)\le S(2)\le\cdots\le S(N)$. By
Step~4, since $S$ decreases at $N{+}1$, we must have $N\ge N^\star$, and
combined with non-decreasing-up-to-$N$, $N = N^\star$. \qed

%% file: tex/appendix/02_algorithm.tex
\section{Algorithm}
\begin{algorithm}[!htbp]
\caption{System-aware Adaptive Tree Construction at Cycle $t$}
\label{alg:adaptive_tree_search}
\DontPrintSemicolon
\KwIn{
Block size $\gamma$; drafter probabilities $q_{t,1:\gamma}$; root/bonus node $r_t$;
context size $c_t$; maximum verification budget $N_{\max}$; draft latency
$T_{\mathrm{draft},t}$; auxiliary-latency $T_{\mathrm{aux},t}$;
AR latency $L_{\mathrm{AR}}$.
}
\KwOut{Selected verification tree $\mathcal T_t^\star$ and budget $N_t^\star$.}

Initialize $\mathcal T \leftarrow \{r_t\}$ and
$\widehat{\bar A} \leftarrow 1$ \tcp*{root/bonus contribution}

Initialize $\mathcal T_t^\star \leftarrow \mathcal T$,
$N_t^\star \leftarrow 0$, and $\widehat S^\star \leftarrow -\infty$\;

Initialize max-heap $H$ with the highest-probability depth-$1$ draft node,
keyed by path score $\rho(\cdot)$\;

\For{$N=1$ \KwTo $N_{\max}$}{
    \If{$H$ is empty}{
        \textbf{break}
    }

    Pop the available node $u$ with largest path score $\rho(u)$ from $H$\;

    $\mathcal T \leftarrow \mathcal T \cup \{u\}$\;
    
    $\widehat{\bar A} \leftarrow \widehat{\bar A} + \rho(u)$\;

    \tcc{Evaluate the estimated speedup after adding node $u$.}
    $\widehat T_{\mathrm{verify}} \leftarrow \textsc{EstimateLatency}(N,c_t)$\;
    
    $\widehat C \leftarrow
    T_{\mathrm{draft},t}
    + \widehat T_{\mathrm{verify}}
    + T_{\mathrm{aux},t}$\;

    $\widehat S \leftarrow \widehat{\bar A}\cdot L_{\mathrm{AR}} / \widehat C$\;

    \If{$\widehat S > \widehat S^\star$}{
        $\widehat S^\star \leftarrow \widehat S$\;
        
        $\mathcal T_t^\star \leftarrow \mathcal T$\;
        
        $N_t^\star \leftarrow N$\;
    }
    \Else {
        \textbf{break}
    }

    \tcc{Update the frontier of the implicit prefix lattice.}
    \If{$\mathrm{depth}(u)<\gamma$}{
        Push the highest-probability child of $u$ into $H$\;
    }
    \If{$u$ has an unpushed next sibling under the same parent}{
        Push the next sibling of $u$ into $H$\;
    }
}

\Return{$\mathcal T_t^\star, N_t^\star$}\;
\end{algorithm}

%% file: tex/appendix/03_overal_pipeline.tex
\section{Overall Pipeline}
\label{app:sec:pipeline}

\begin{figure}[h]
  \centering
  \includegraphics[width=\textwidth]{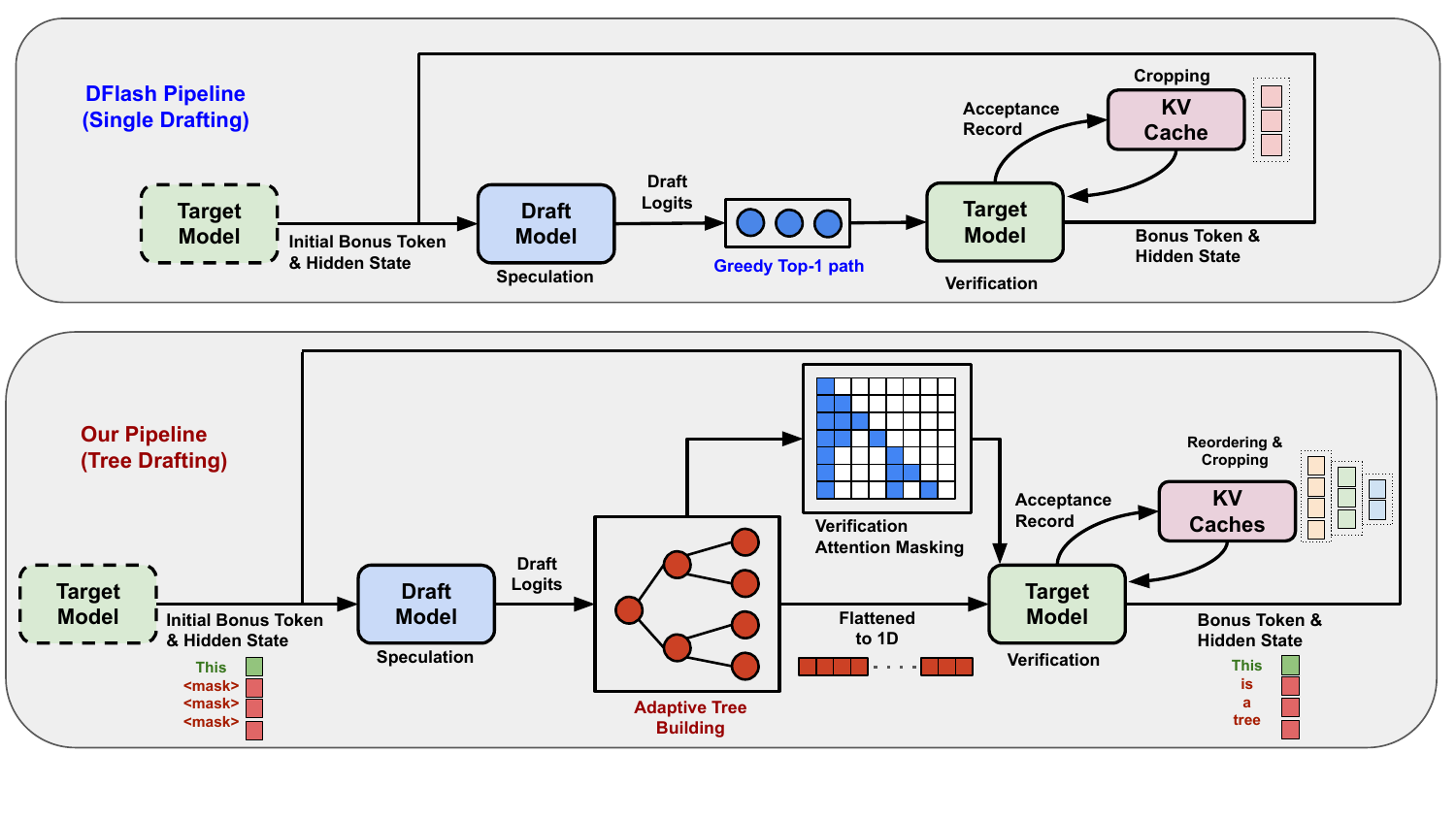}
  \caption{\textbf{Single-path vs.\ tree-based speculative decoding.}
  \emph{Top:} the DFlash baseline drafts a single greedy top-1 continuation, which the target verifies in one forward pass before cropping the KV cache to the accepted prefix.
  \emph{Bottom:} our tree pipeline. From the same initial bonus token and hidden state, the draft model emits per-position logits that the adaptive tree builder expands into a token tree. The tree is linearized to a 1D input for the target, and a custom verification attention mask preserves parent--child dependencies in a single batched forward pass. The acceptance record tracks the path obtained by matching each child to the target's next token at the visited prefix; the target's KV cache is then reordered and cropped to retain only accepted nodes, and the new bonus token and hidden state feed the next iteration.}
  \label{fig:pipeline}
\end{figure}

Figure~\ref{fig:pipeline} (bottom) summarizes one iteration of our pipeline; we walk through its four stages below.

\paragraph{Drafting.}
Each decoding step begins from the bonus token and hidden state produced by the previous target forward pass, or, on the first step, from a short prefill. The draft model consumes these and emits a distribution over candidate continuations at multiple future positions. Rather than committing to a single greedy path as in the baseline (Figure~\ref{fig:pipeline}, top), we feed these draft distributions into an adaptive tree builder that grows a token tree under a fixed verification budget, allocating depth and width to the most promising branches.

\paragraph{Linearization and verification mask.}
The tree is linearized into a flat sequence for the target model. To make a single batched forward pass behave as if each branch were verified independently, we construct a custom attention mask (Figure~\ref{fig:pipeline}, center-bottom): every tree node attends to the shared prefix in the KV cache and to its own ancestors in the tree, but not to siblings or unrelated branches. Position ids are derived from each node's depth, so rotary embeddings see the correct relative position regardless of the linearization order.

\paragraph{Acceptance.}
After the target forward pass, we descend the tree from the root, drawing the target's next token at each visited prefix and following the child that matches it. The descent stops at the first drawn token the tree does not contain, which is also emitted. The accepted nodes' tokens are emitted, and the target's KV cache is reordered and cropped so that it contains the shared prefix followed only by the accepted path, discarding the rest of the tree.

\paragraph{Iteration.}
The hidden state of the deepest accepted node, together with the target's next-token sample at that position, becomes the bonus token and hidden state that seed the next iteration. The cycle repeats until an EOS token or length limit is reached.

%% file: tex/appendix/04_additional_exps.tex
\section{Additional Experimental Results}
\label{app:sec:add_exp_ret}

\subsection{Path Score Validation}
\label{app:top1_path_score_validation}

\begin{figure}[!htbp]
    \centering
    \includegraphics[width=0.6\linewidth]{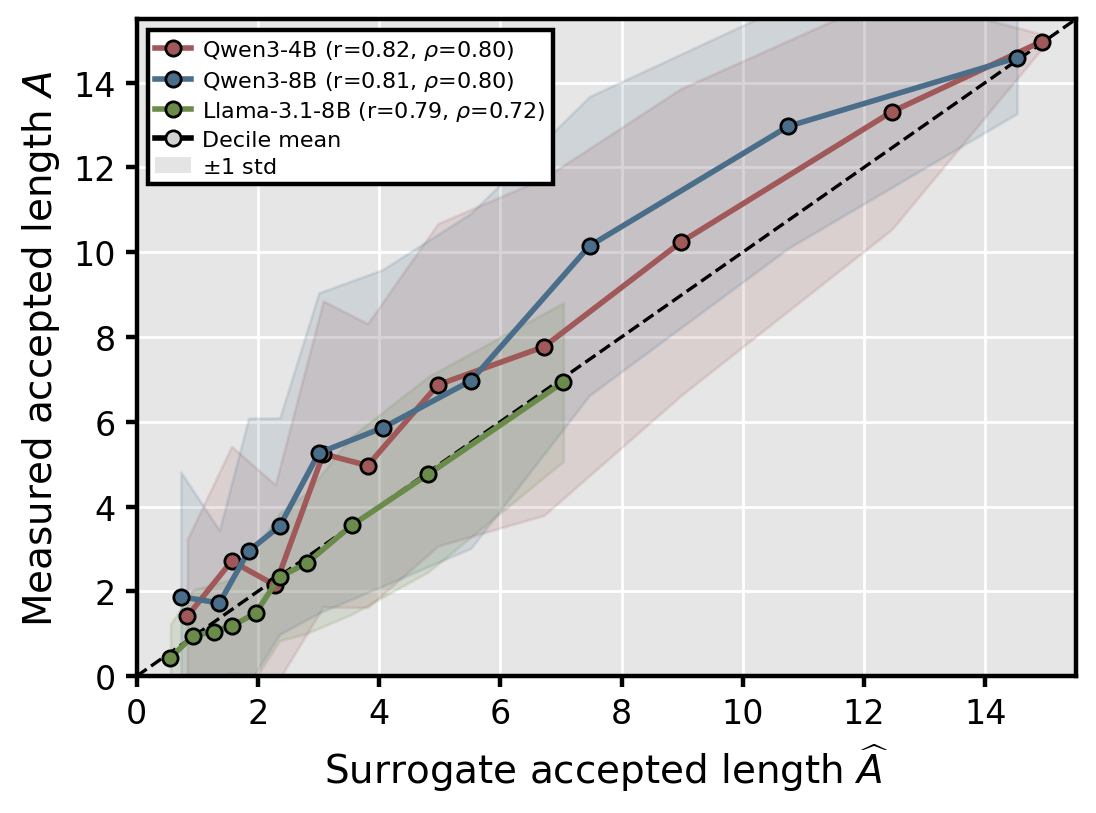}
    \caption{\textbf{Path score validation.}
    For each decode step we collapse the draft tree to a single greedy path by taking the drafter's top-1 token $x_k=\arg\max_v q_k(v)$ at every position $k\in\{1,\ldots,\gamma\}$, where $\gamma$ is the block size and $q_k(v)$ is the drafter distribution of vocabulary $v$ for position $k$. Then we evaluate the surrogate accepted length specialized to this path,
    $\widehat{A}=\sum_{k=1}^{\gamma}\prod_{j=1}^{k} q_j(x_j)$,
    which is the tree sum of path scores $\rho(i)$ from~\eqref{eq:path-score} restricted to a depth-$\gamma$ chain (root term omitted to match the measured length below). Setup: DFlash block-diffusion drafters paired with each target, evaluated on MATH500 with $20$ prompts at temperature $0$; block size is $16$ for Qwen and $10$ for Llama. $\widehat{A}$ correlates strongly with the target-verified accepted length $A$ across all three target/draft pairs (Pearson $0.79$--$0.82$, Spearman $0.72$--$0.80$), supporting the use of drafter path probabilities as a proxy for target acceptance in Section~\ref{sec:method:surrogate}.}
    \label{fig:mean_al_correlate}
\end{figure}

\subsection{Additional Experimental on H100}
\label{app:subsec:h100}

\input{tables/appendix/main_experiment_h100}

\subsection{Comparison with Concurrent Work: DDTree}
\label{app:subsec:ddtree}

\paragraph{Concurrent and independent development.}
As discussed in Section~\ref{sec:related_work}, DDTree~\citep{ringel2026accelerating}
appeared concurrently with our work, after the release of
DFlash~\citep{chen2026dflash}, when the core ideas of \algo were largely
finalized. We developed our method, including best-first tree expansion,
entirely independently of DDTree, starting from our own findings. The two
approaches nevertheless arrived at equivalent fixed-budget tree construction
under the same drafter path-score surrogate. Best-first search, like
breadth-first search, is a classical search technique in computer science.
Moreover, using drafter confidence and path-probability scores as proxies for
target-model acceptance has precedents in tree-based speculative decoding,
including the EAGLE family, particularly
EAGLE-2~\citep{li2024eagle,li2024eagle2fasterinferencelanguage}.
These shared foundations make independent convergence on a similar tree
construction method plausible. We therefore directly compare DDTree with
\algo to examine how this theoretical equivalence translates across separate
implementations and to assess the additional benefit of adaptive budget control.

\paragraph{Experimental setup.}
We evaluate Qwen3-8B and Llama-3.1-8B-Instruct on an NVIDIA A100 PCIe GPU with
40\,GB of memory, using greedy decoding at
temperature $T=0$. We use the same eight math, code, and chat benchmarks and
evaluation settings as in our main experiments, and report benchmark averages.
The DDTree results are obtained by running code downloaded from its official
GitHub repository.\footnote{\url{https://github.com/liranringel/ddtree}}
We compare DDTree and \algo-Fixed at matched budgets
$N\in\{32,64,128,256,512,1024\}$, measuring speedup over target-only
autoregressive decoding and AAL ($\tau$).

\input{tables/appendix/ddtree_fixed_budget}

\paragraph{Fixed-budget comparison.}
\autoref{tab:ddtree_fixed_budget} shows closely matching AAL:
the absolute difference in $\tau$ is at most $0.04$ tokens across all tested
model--budget pairs. Speedups are also close at most budgets, and both methods
attain their highest average speedup at $N=128$ for Qwen3-8B and $N=64$ for
Llama-3.1-8B-Instruct. At $N=256$, however, \algo-Fixed is approximately
$10\%$ slower than DDTree despite similar AAL.
Overall, these results provide empirical evidence consistent with the
theoretical equivalence of the fixed-budget best-first construction across
independent implementations, while showing that equivalent tree-selection
rules need not yield identical runtime performance.

\paragraph{Adaptive budget control without a budget sweep.}
\autoref{tab:ddtree_adaptive} compares adaptive \algo with the best fixed-budget
speedup of each method, selected in hindsight from the sweep above. Without a
separate budget sweep, \algo achieves $5.78\times$ speedup on Qwen3-8B and
$4.58\times$ on Llama-3.1-8B-Instruct, recovering $92.6\%$ and $99.3\%$,
respectively, of DDTree's best fixed-budget speedup. Thus, \algo recovers at
least $92.6\%$ of the best DDTree result across these two targets through
online budget selection, without the manual budget tuning used to select
DDTree's best fixed configuration.

\input{tables/appendix/ddtree_adaptive}

\clearpage
\subsection{Experiments with MoE Architectures}
\label{app:subsec:moe}

We extend our evaluation to Qwen3-Coder-30B-A3B-Instruct, a mixture-of-experts
(MoE) target model, using its compatible block-diffusion drafter from
DFlash~\citep{chen2026dflash}. Since this target specializes in code generation,
we evaluate on three coding benchmarks: HumanEval~\citep{chen2021evaluating},
LiveCodeBench~\citep{jain2024livecodebench}, and MBPP~\citep{austin2021program}.
All MoE experiments are conducted on an NVIDIA A100 GPU with 80\,GB of memory.
For simplicity, we restrict these experiments to greedy decoding at temperature
$T=0$. We report speedup over target-only autoregressive decoding and AAL
($\tau$) across these benchmarks.

\paragraph{Fixed-budget sweep.}
We first examine the behavior of best-first tree expansion with \algo-Fixed,
sweeping the verification budget from $N=16$ to $N=1025$.
\autoref{tab:moe_fixed_budget} shows that increasing the budget raises $\tau$
from $8.22$ to $10.60$, while speedup decreases from $2.07\times$ to
$1.37\times$. Thus, the additional accepted tokens from larger trees do not
translate into higher speedup in this setting; among the tested budgets,
$N=16$ gives the highest speedup.

\input{tables/appendix/moe_fixed_budget}

\paragraph{Comparison with DFlash.}
We next compare adaptive \algo against DFlash with a drafting length of $16$.
We do not include EAGLE-3 because it does not provide a compatible drafter for
Qwen3-Coder-30B-A3B-Instruct, preventing a comparison on this target.
As shown in \autoref{tab:moe_comparison}, \algo achieves a speedup of
$1.93\times$ and $\tau=8.56$, compared with $1.71\times$ and $\tau=7.43$ for
DFlash. These results extend the observed gains in speedup and AAL to an MoE
target model.

\input{tables/appendix/moe_comparison}

\clearpage
\subsection{Fair Comparison with EAGLE-3}
\label{app:subsec:eagle3_fairness}

\paragraph{Motivation and setup.}
Our main experiments use a default tree budget of $60$ for EAGLE-3, whereas
\algo explores different budgets through fixed-budget sweeps or adaptive
budget control. This difference raises the question of whether the observed
gains reflect a more favorable budget choice for \algo. To examine comparison
fairness, we tune EAGLE-3's tree-construction hyperparameters and compare it
with \algo-Fixed at identical tree budgets. We evaluate Qwen3-8B on an NVIDIA
A100 GPU with 40\,GB of memory using greedy decoding at temperature $T=0$.
For simplicity, we use two representative benchmarks, MATH500 and HumanEval,
and report averages over these benchmarks.

\paragraph{EAGLE-3 hyperparameter tuning.}
EAGLE-3~\citep{li2025eagle} uses an autoregressive drafter with beam-search-style
tree expansion. We vary the tree depth $d$ and the number $k$ of top-ranked
child candidates considered per node at each expansion depth. For a total
tree budget $N$, the expansion must supply $N-1$ nodes in addition to the
root/bonus token, giving the candidate-count constraint
\[
    k + (d-1) k^2 \geq N-1.
\]
This constraint gives a lower bound on $d$ for candidate-count feasibility;
it does not identify the depth that maximizes speedup. Increasing $k$ explores
more child candidates at each depth, while increasing $d$ raises the maximum
possible acceptance length. Both can improve the candidate tree, but incur
additional search and drafting costs. Width expansion can process candidates
in parallel as batched sequences within the autoregressive drafter, whereas
each additional depth requires another sequential autoregressive step.
Consequently, the cost of deeper drafting can outweigh the benefit of finding
better candidate paths. We cap $d$ at $12$ to limit this sequential cost.

We sweep $k\in\{5,10,15\}$. For each $(N,k)$, we choose a reference depth
in advance, guided by the default EAGLE-3 settings, to balance sufficient
drafting length against autoregressive drafting cost. We then test neighboring
depths with offsets $\{-1,0,+1\}$, retaining only configurations satisfying
the candidate-count constraint and $d\leq12$. Thus, the reference depth is
not simply the minimum feasible depth, and this is a local hyperparameter
sweep rather than an exhaustive search over all feasible depths.
At each budget, we report the
best tested EAGLE-3 configuration by average speedup. \algo-Fixed uses the
DFlash block-diffusion drafter with best-first tree expansion and has no
additional tree-construction hyperparameters to tune beyond $N$.

\paragraph{Comparison across matched tree budgets.}
At the same tree budget of $N=60$, \algo-Fixed achieves $2.53\times$ the
speedup and $2.55\times$ the AAL of the best tested
EAGLE-3 configuration as shown in \autoref{tab:eagle3_matched_60}.

\input{tables/appendix/eagle3_matched_60}

\clearpage
We extend this comparison to $N\in\{33,65,129,257,513,1025\}$, applying the same EAGLE-3
hyperparameter-tuning policy at each budget.
\autoref{tab:eagle3_budget_sweep} shows that \algo-Fixed achieves
$2.52$--$2.88\times$ the speedup of the best tested EAGLE-3 configuration
across all matched budgets. These comparisons provide empirical evidence
that the gains observed in the main experiments cannot be explained solely
by allocating a larger tree budget to \algo: best-first tree expansion with
the DFlash drafter remains faster under identical budget constraints.

\input{tables/appendix/eagle3_budget_sweep}

\paragraph{Adaptive budget control.}
Without a budget sweep, adaptive \algo achieves $7.61\times$ speedup with
an average tree size of $157.1$, recovering $93.0\%$ of the best
\algo-Fixed speedup ($8.18\times$ at $N=129$). It also achieves
$2.35\times$ the speedup of the best EAGLE-3 configuration across the tested
budgets ($3.24\times$ at $N=129$, $(d,k)=(4,15)$). This adaptive comparison
uses a varying tree budget; the fixed-budget comparisons above establish
the advantage under matched budgets.

%% file: tables/appendix/main_experiment_h100.tex
\begin{table*}[!htbp]
\centering
\caption{\textbf{Additional results on H100.} Speedup and AAL ($\tau$) for Qwen3-4B (Q3-4B) and Qwen3-8B (Q3-8B) under greedy decoding ($T=0$). For the baselines, the tree budget size of EAGLE-3 is set to 60. The block size of DFlash is 16 for both Qwen3-4B and Qwen3-8B. \textbf{Bold} indicates the highest value within each (model, benchmark) pair.}
\label{tab:add_res_h100}
\setlength{\tabcolsep}{2pt}
\renewcommand{\arraystretch}{1.05}
\resizebox{\textwidth}{!}{%
\begin{tabular}{@{}c c *{9}{cc}@{}}
\toprule
& & \multicolumn{6}{c}{\textbf{Math}} & \multicolumn{6}{c}{\textbf{Code}} & \multicolumn{4}{c}{\textbf{Chat}} & \multicolumn{2}{c}{} \\
\cmidrule(lr){3-8} \cmidrule(lr){9-14} \cmidrule(lr){15-18}
\textbf{Model} & \textbf{Method}
& \multicolumn{2}{c}{GSM8K} & \multicolumn{2}{c}{MATH500} & \multicolumn{2}{c}{AIME25}
& \multicolumn{2}{c}{HumanEval} & \multicolumn{2}{c}{MBPP} & \multicolumn{2}{c}{LCB}
& \multicolumn{2}{c}{MT-Bench} & \multicolumn{2}{c}{Alpaca}
& \multicolumn{2}{c}{\textbf{Average}} \\
\cmidrule(lr){3-4}\cmidrule(lr){5-6}\cmidrule(lr){7-8}\cmidrule(lr){9-10}\cmidrule(lr){11-12}\cmidrule(lr){13-14}\cmidrule(lr){15-16}\cmidrule(lr){17-18}\cmidrule(lr){19-20}
& & Speedup & $\tau$ & Speedup & $\tau$ & Speedup & $\tau$ & Speedup & $\tau$ & Speedup & $\tau$ & Speedup & $\tau$ & Speedup & $\tau$ & Speedup & $\tau$ & Speedup & $\tau$ \\
\midrule
 & EAGLE-3 & $3.23\times$ & 2.80 & $3.04\times$ & 2.54 & $3.13\times$ & 2.51 & $3.29\times$ & 2.54 & $2.95\times$ & 2.47 & $2.86\times$ & 2.24 & $2.89\times$ & 2.49 & $2.55\times$ & 2.17 & $2.99\times$ & 2.47 \\
 & DFlash & $5.83\times$ & 6.51 & $6.96\times$ & 7.79 & $6.45\times$ & 7.26 & $5.87\times$ & 6.69 & $5.40\times$ & 6.12 & $5.98\times$ & 6.97 & $3.21\times$ & 4.38 & $2.49\times$ & 3.09 & $5.27\times$ & 6.10 \\
\rowcolor{blue!10} \cellcolor{white} \multirow{-3}{*}{Q3-4B} & \textbf{\algo} & $\mathbf{7.26\times}$ & $\mathbf{8.91}$ & $\mathbf{8.05\times}$ & $\mathbf{10.20}$ & $\mathbf{7.96\times}$ & $\mathbf{9.43}$ & $\mathbf{8.13\times}$ & $\mathbf{9.23}$ & $\mathbf{7.22\times}$ & $\mathbf{8.81}$ & $\mathbf{8.37\times}$ & $\mathbf{9.36}$ & $\mathbf{4.47\times}$ & $\mathbf{6.34}$ & $\mathbf{3.58\times}$ & $\mathbf{4.73}$ & $\mathbf{6.88\times}$ & $\mathbf{8.38}$ \\
\midrule
 & EAGLE-3 & $3.16\times$ & 2.69 & $2.83\times$ & 2.48 & $2.85\times$ & 2.50 & $2.43\times$ & 2.65 & $2.82\times$ & 2.20 & $2.17\times$ & 1.81 & $2.49\times$ & 2.25 & $2.05\times$ & 1.73 & $2.60\times$ & 2.29 \\
 & DFlash & $5.87\times$ & 6.55 & $6.86\times$ & 7.93 & $5.95\times$ & 7.18 & $5.83\times$ & 6.56 & $5.27\times$ & 5.98 & $6.04\times$ & 7.16 & $3.03\times$ & 4.23 & $2.47\times$ & 3.09 & $5.17\times$ & 6.09 \\
\rowcolor{blue!10} \cellcolor{white} \multirow{-3}{*}{Q3-8B} & \textbf{\algo} & $\mathbf{7.64\times}$ & $\mathbf{9.02}$ & $\mathbf{8.02\times}$ & $\mathbf{10.21}$ & $\mathbf{6.88\times}$ & $\mathbf{9.35}$ & $\mathbf{7.33\times}$ & $\mathbf{9.20}$ & $\mathbf{7.04\times}$ & $\mathbf{8.62}$ & $\mathbf{7.30\times}$ & $\mathbf{9.48}$ & $\mathbf{4.12\times}$ & $\mathbf{6.16}$ & $\mathbf{3.35\times}$ & $\mathbf{4.82}$ & $\mathbf{6.46\times}$ & $\mathbf{8.36}$ \\
\bottomrule
\end{tabular}}
\end{table*}

%% file: tables/appendix/ddtree_fixed_budget.tex
\begin{table}[!htbp]
\centering
\captionsetup{position=top,skip=7pt}
\caption{\textbf{Fixed-budget comparison with DDTree.} Average speedup over autoregressive decoding and AAL ($\tau$) across the eight main math, code, and chat benchmarks on an NVIDIA A100 PCIe 40\,GB GPU at $T=0$. Each budget has separate Speedup and $\tau$ columns. Q3-8B and L3.1-8B denote Qwen3-8B and Llama-3.1-8B-Instruct, respectively. \textbf{Bold} indicates the highest speedup across budgets within each row.}
\label{tab:ddtree_fixed_budget}
\small
\setlength{\tabcolsep}{3pt}
\renewcommand{\arraystretch}{1.1}
\resizebox{\textwidth}{!}{%
\begin{tabular}{@{}ll *{6}{cc}@{}}
\toprule
\multirow{2}{*}{\textbf{Target}} & \multirow{2}{*}{\textbf{Method}}
& \multicolumn{2}{c}{$N=32$} & \multicolumn{2}{c}{$N=64$}
& \multicolumn{2}{c}{$N=128$} & \multicolumn{2}{c}{$N=256$}
& \multicolumn{2}{c}{$N=512$} & \multicolumn{2}{c}{$N=1024$} \\
\cmidrule(lr){3-4}\cmidrule(lr){5-6}\cmidrule(lr){7-8}\cmidrule(lr){9-10}\cmidrule(lr){11-12}\cmidrule(lr){13-14}
& & Speedup & $\tau$ & Speedup & $\tau$ & Speedup & $\tau$ & Speedup & $\tau$ & Speedup & $\tau$ & Speedup & $\tau$ \\
\midrule
\multirow{2}{*}{Q3-8B} & DDTree & $5.64\times$ & 7.52 & $5.91\times$ & 7.98 & $\mathbf{6.24\times}$ & $\mathbf{8.31}$ & $5.53\times$ & 8.67 & $3.84\times$ & 8.98 & $2.36\times$ & 9.15 \\
& \algo-Fixed & $5.73\times$ & 7.55 & $6.03\times$ & 7.99 & $\mathbf{6.37\times}$ & $\mathbf{8.35}$ & $4.97\times$ & 8.65 & $3.82\times$ & 9.00 & $2.35\times$ & 9.16 \\
\midrule
\multirow{2}{*}{L3.1-8B} & DDTree & $4.40\times$ & 5.52 & $\mathbf{4.61\times}$ & $\mathbf{5.84}$ & $4.24\times$ & 6.10 & $3.64\times$ & 6.30 & $2.46\times$ & 6.52 & $1.49\times$ & 6.76 \\
& \algo-Fixed & $4.49\times$ & 5.54 & $\mathbf{4.66\times}$ & $\mathbf{5.80}$ & $4.27\times$ & 6.07 & $3.27\times$ & 6.33 & $2.46\times$ & 6.51 & $1.49\times$ & 6.74 \\
\bottomrule
\end{tabular}}
\end{table}

%% file: tables/appendix/ddtree_adaptive.tex
\begin{table}[!htbp]
\centering
\captionsetup{position=top,skip=7pt}
\caption{\textbf{Recovery of DDTree's best fixed-budget speedup without a budget sweep.} Same setup as \autoref{tab:ddtree_fixed_budget}. The best fixed results are selected by maximum benchmark-average speedup over the tested budgets; parentheses give the selected budget. \algo uses adaptive budget control. The final column is adaptive \algo speedup divided by DDTree's best fixed-budget speedup.}
\label{tab:ddtree_adaptive}
\small
\setlength{\tabcolsep}{5pt}
\renewcommand{\arraystretch}{1.1}
\resizebox{\textwidth}{!}{%
\begin{tabular}{@{}lcccc@{}}
\toprule
\textbf{Target} & \textbf{DDTree best fixed} & \textbf{\algo-Fixed best} & \textbf{\algo} & \textbf{\algo{} / DDTree best fixed} \\
\midrule
Qwen3-8B & $6.24\times$ ($N=128$) & $6.37\times$ ($N=128$) & $5.78\times$ & $\mathbf{92.6\%}$ \\
Llama-3.1-8B-Instruct & $4.61\times$ ($N=64$) & $4.66\times$ ($N=64$) & $4.58\times$ & $\mathbf{99.3\%}$ \\
\bottomrule
\end{tabular}}
\end{table}

%% file: tables/appendix/moe_fixed_budget.tex
\begin{table}[!htbp]
\centering
\captionsetup{position=top,skip=7pt}
\caption{\textbf{Fixed-budget sweep on Qwen3-Coder-30B-A3B-Instruct.} Speedup over target-only autoregressive decoding and AAL ($\tau$) for \algo-Fixed with best-first tree expansion on HumanEval, LiveCodeBench, and MBPP at $T=0$. \textbf{Bold} indicates the highest value in each metric column.}
\label{tab:moe_fixed_budget}
\small
\setlength{\tabcolsep}{12pt}
\renewcommand{\arraystretch}{1.05}
\begin{tabular}{rcc}
\toprule
\textbf{Budget $N$} & \textbf{Speedup} & \textbf{$\tau$} \\
\midrule
16   & $\mathbf{2.07\times}$ & $8.22$ \\
33   & $1.94\times$ & $8.96$ \\
65   & $1.74\times$ & $9.23$ \\
129  & $1.66\times$ & $9.76$ \\
257  & $1.60\times$ & $10.13$ \\
513  & $1.46\times$ & $10.28$ \\
1025 & $1.37\times$ & $\mathbf{10.60}$ \\
\bottomrule
\end{tabular}
\end{table}

%% file: tables/appendix/moe_comparison.tex
\begin{table}[!htbp]
\centering
\captionsetup{position=top,skip=7pt}
\caption{\textbf{Comparison with DFlash on Qwen3-Coder-30B-A3B-Instruct.} Speedup over target-only autoregressive decoding and AAL ($\tau$) on HumanEval, LiveCodeBench, and MBPP at $T=0$. DFlash uses a drafting length of $16$, and \algo uses adaptive budget selection. \textbf{Bold} indicates the highest value in each metric column.}
\label{tab:moe_comparison}
\small
\setlength{\tabcolsep}{12pt}
\renewcommand{\arraystretch}{1.05}
\begin{tabular}{lcc}
\toprule
\textbf{Method} & \textbf{Speedup} & \textbf{$\tau$} \\
\midrule
DFlash & $1.71\times$ & $7.43$ \\
\rowcolor{blue!10}
\algo & $\mathbf{1.93\times}$ & $\mathbf{8.56}$ \\
\bottomrule
\end{tabular}
\end{table}

%% file: tables/appendix/eagle3_matched_60.tex
\begin{table}[!htbp]
\centering
\captionsetup{position=top,skip=7pt}
\caption{\textbf{Comparison with tuned EAGLE-3 at a matched tree budget of $N=60$.} Average speedup over autoregressive decoding and AAL ($\tau$) on MATH500 and HumanEval using Qwen3-8B on an NVIDIA A100 40\,GB GPU at $T=0$. EAGLE-3 uses the configuration with the highest average speedup in the tested search space. \textbf{Bold} indicates the higher value in each metric column.}
\label{tab:eagle3_matched_60}
\small
\setlength{\tabcolsep}{10pt}
\renewcommand{\arraystretch}{1.1}
\begin{tabular}{@{}lccc@{}}
\toprule
\textbf{Method} & $N$ & \textbf{Speedup} & $\tau$ \\
\midrule
EAGLE-3, best tested $(d,k)=(5,15)$ & 60 & $3.06\times$ & 3.66 \\
\algo-Fixed & 60 & $\mathbf{7.75\times}$ & $\mathbf{9.33}$ \\
\bottomrule
\end{tabular}
\end{table}

%% file: tables/appendix/eagle3_budget_sweep.tex
\begin{table}[!htbp]
\centering
\captionsetup{position=top,skip=7pt}
\caption{\textbf{Budget scaling with EAGLE-3 tuned at each matched budget.} Same setup as \autoref{tab:eagle3_matched_60}. Speedups are relative to autoregressive decoding. The best tested EAGLE-3 $(d,k)$ is selected separately at each $N$; $(5,10)$ and $(5,15)$ tie at $N=257$. Ratio denotes \algo-Fixed speedup divided by EAGLE-3 speedup. \textbf{Bold} indicates the higher speedup within each row.}
\label{tab:eagle3_budget_sweep}
\small
\setlength{\tabcolsep}{8pt}
\renewcommand{\arraystretch}{1.1}
\begin{tabular}{@{}rcccc@{}}
\toprule
$N$ & \textbf{Best EAGLE-3 $(d,k)$} & \textbf{EAGLE-3} & \textbf{\algo-Fixed} & \textbf{Ratio} \\
\midrule
33 & $(4,10)$ & $2.97\times$ & $\mathbf{7.53\times}$ & $2.54\times$ \\
65 & $(4,15)$ & $3.12\times$ & $\mathbf{7.91\times}$ & $2.54\times$ \\
129 & $(4,15)$ & $3.24\times$ & $\mathbf{8.18\times}$ & $2.52\times$ \\
257 & $(5,10)$, $(5,15)$ & $2.20\times$ & $\mathbf{6.33\times}$ & $2.88\times$ \\
513 & $(5,15)$ & $1.83\times$ & $\mathbf{4.82\times}$ & $2.63\times$ \\
1025 & $(6,15)$ & $1.10\times$ & $\mathbf{3.00\times}$ & $2.73\times$ \\
\bottomrule
\end{tabular}
\end{table}

%% file: tex/appendix/05_cost_models.tex
\section{Analytical Cost Modeling}
\label{app:sec:cost_model}

We estimate verifier latency with a roofline-style predictor,
\[
R_t(N)=\max\!\left(
\frac{\mathrm{FLOPs}(N,C_t)}{\mathrm{PeakFLOPs}},
\frac{\mathrm{Bytes}(N,C_t)}{\mathrm{Bandwidth}}
\right),
\]
where $s=N$ is the verification budget and $c=C_t$ is the cached context length.
We derive the FLOPs and memory traffic for dense Transformers and the
Qwen3-Coder-30B-A3B-Instruct MoE architecture evaluated in this work. We assume non-fused
SDPA with materialized attention scores and count one multiply-accumulate as
2 FLOPs. Normalization, positional encoding, nonlinearities, biases, router
softmax/top-$K$ selection, and expert dispatch/combination overheads are omitted
from this matrix-multiplication and tensor-I/O approximation.

\paragraph{Scope of the MoE model.}
The MoE derivation is specific to \textbf{Qwen3-Coder-30B-A3B-Instruct},
the target evaluated in Appendix~\ref{app:subsec:moe}. This architecture uses
full attention in every layer and an FFN modeled as a router and $K$ active
gated experts per token, without an additional shared-expert branch.
Other MoE architectures require separate derivations reflecting their own
attention, expert, and routing designs.

\paragraph{Notation.}
Table~\ref{tab:cost_notation} defines the dimensions. We write
$h_q=n_qd$ and $h_{kv}=n_{kv}d$. For dense models, $h_{ffn}$ is the dense
FFN width; for Qwen MoE, it is the width of \emph{one expert}.

\begin{table}[!htbp]
\centering
\captionsetup{position=top,skip=7pt}
\caption{Notation used in the cost model.}
\label{tab:cost_notation}
\small
\begin{tabular}{ll}
\toprule
\textbf{Symbol} & \textbf{Meaning} \\
\midrule
$s, c$ & newly verified tokens and cached context length \\
$L, h$ & number of Transformer layers and hidden size \\
$n_q, n_{kv}, d$ & query heads, key/value heads, and head dimension \\
$h_{ffn}, V$ & dense/per-expert FFN width and vocabulary size \\
$bp$ & bytes per parameter/activation element ($2$ in the implementation) \\
$E, K$ & (MoE) total experts per layer and selected experts per token \\
$U(s)$ & (MoE) modeled distinct experts accessed per layer \\
\bottomrule
\end{tabular}
\end{table}

\subsection{FLOPs}
\label{app:subsec:cost_flops}

Table~\ref{tab:flops_breakdown} lists the dominant FLOP terms for dense
Transformers and Qwen MoE.

\begin{table}[H]
\centering
\captionsetup{position=top,skip=7pt}
\caption{FLOP breakdown. Entries are per layer unless marked otherwise.
Dense and Qwen MoE models share the full-attention terms.
The Qwen MoE FFN replaces the dense FFN.}
\label{tab:flops_breakdown}
\small
\setlength{\tabcolsep}{4pt}
\begin{tabular}{lll}
\toprule
\textbf{Module} & \textbf{Operation} & \textbf{FLOPs} \\
\midrule
\multirow{5}{*}{Attention} & $Q$ projection & $2shh_q$ \\
 & $K,V$ projections & $4shh_{kv}$ \\
 & attention scores $QK^\top$ & $2s(c+s)h_q$ \\
 & attention-value product & $2s(c+s)h_q$ \\
 & output projection & $2sh_qh$ \\
\midrule
\textbf{Attention total} & all layers & $L[4shh_q+4shh_{kv}+4s(c+s)h_q]$ \\
\midrule
\multirow{2}{*}{Dense FFN} & gate/up projections & $4shh_{ffn}$ \\
 & down projection & $2sh_{ffn}h$ \\
\textbf{Dense FFN total} & all layers & $6Lshh_{ffn}$ \\
\midrule
\multirow{3}{*}{MoE FFN} & router projection & $2shE$ \\
 & selected experts: gate/up & $4sKhh_{ffn}$ \\
 & selected experts: down & $2sKh_{ffn}h$ \\
\textbf{MoE FFN total} & all layers & $L(2shE+6sKhh_{ffn})$ \\
\midrule
\textbf{LM head} & hidden $\rightarrow$ vocab & $2shV$ \\
\bottomrule
\end{tabular}
\end{table}

Summing the attention, FFN, and output-head terms gives the following
architecture-specific totals (each uses its own model dimensions):
\begin{equation}
\label{eq:total_flops_appendix}
\begin{aligned}
\mathrm{FLOPs}_{\mathrm{dense}}
&=L\big[4shh_q+4shh_{kv}+4s(c+s)h_q+6shh_{ffn}\big]+2shV,\\
\mathrm{FLOPs}_{\mathrm{Qwen\ MoE}}
&=L\big[4shh_q+4shh_{kv}+4s(c+s)h_q+2shE+6sKhh_{ffn}\big]\\
&\quad+2shV.
\end{aligned}
\end{equation}

\subsection{Memory Traffic}
\label{app:subsec:cost_memory}

We decompose memory traffic into (i) model-weight reads, (ii) KV-cache
reads/writes, and (iii) intermediate activation I/O. These terms approximate
bytes transferred per verification pass, rather than the total resident model
size or peak allocated memory. Table~\ref{tab:memory_breakdown} gives the
common terms and the Qwen MoE replacements. Dispatch buffers and
output-combination traffic are omitted.

\paragraph{Expert-weight reads.}
The Qwen MoE implementation uses
\[
U(s)=\min(E,sK),\qquad
W_{\mathrm{ffn}}^{\mathrm{Qwen}}=hE+3U(s)hh_{ffn}
\]
as its per-layer FFN weight-read count in elements. There are $sK$
token--expert assignments but at most $E$ distinct experts. The expression
$U(s)$ is an upper bound on the number of distinct experts touched, rather
than an expectation over routing decisions; each touched expert's weights
are counted once per pass. The estimate saturates at all $E$ experts when
$s\geq E/K$.

\begin{table}[H]
\centering
\captionsetup{position=top,skip=7pt}
\caption{Memory-traffic breakdown in elements; multiply by $bp$ for bytes.
$W_{\mathrm{ffn}}$ and $I_{\mathrm{ffn}}$ select the dense or Qwen MoE FFN
terms; for MoE, $I_{\mathrm{ffn}}$ sums the router and expert-projection rows.
Attention and KV-cache terms are shared by both architectures.}
\label{tab:memory_breakdown}
\small
\setlength{\tabcolsep}{3pt}
\begin{tabular}{lll}
\toprule
\textbf{Category} & \textbf{Module} & \textbf{Read / Write} \\
\midrule
\multirow{4}{*}{Weights} & embedding + LM head & $2Vh$ \\
 & attention / layer & $2h(h_q+h_{kv})$ \\
 & dense FFN / layer & $W_{\mathrm{ffn}}=3hh_{ffn}$ \\
 & Qwen MoE FFN / layer & $W_{\mathrm{ffn}}=hE+3U(s)hh_{ffn}$ \\
\textbf{Weights total} & all layers + head & $2Vh+L[2h(h_q+h_{kv})+W_{\mathrm{ffn}}]$ \\
\midrule
\multirow{2}{*}{KV cache} & past-context read / layer & $2ch_{kv}$ \\
 & new-token write / layer & $2sh_{kv}$ \\
\textbf{KV cache total} & all layers & $2Lh_{kv}(c+s)$ \\
\midrule
\multirow{5}{*}{Activations} & attention / layer & $2sh+4sh_q+2sh_{kv}+2n_qs(c+s)$ \\
 & dense FFN / layer & $I_{\mathrm{ffn}}=2sh+4sh_{ffn}$ \\
 & MoE router / layer & $sh+sE$ \\
 & MoE expert projections / layer & $2sh+4sKh_{ffn}$ \\
 & LM head & $s(h+V)$ \\
\textbf{Activations total} & all layers + head & $\begin{aligned}
&L[2sh+4sh_q+2sh_{kv}+I_{\mathrm{ffn}}]\\
&\quad+2Ln_qs(c+s)+s(h+V)
\end{aligned}$ \\
\bottomrule
\end{tabular}
\end{table}

Adding the three categories yields the dense and Qwen MoE totals:
\begin{equation}
\label{eq:total_bytes_appendix}
\begin{aligned}
\mathrm{Bytes}_{\mathrm{dense}}=bp\Big\{&2Vh+s(h+V)\\
&+L\big[2h(h_q+h_{kv})+3hh_{ffn}+2h_{kv}(c+2s)\\
&\qquad+4s(h+h_q+h_{ffn})+2n_qs(c+s)\big]\Big\},\\
\mathrm{Bytes}_{\mathrm{Qwen\ MoE}}=bp\Big\{&2Vh+s(h+V)\\
&+L\big[2h(h_q+h_{kv})+hE+3U(s)hh_{ffn}\\
&\qquad+2h_{kv}(c+2s)+s(5h+4h_q+4Kh_{ffn}+E)\\
&\qquad+2n_qs(c+s)\big]\Big\}.
\end{aligned}
\end{equation}

\paragraph{Precision and implementation assumptions.}
Even when input embeddings and the LM head are tied, their weight accesses
are counted separately. The implementation uses $bp=2$ for both weights and
activations.
The formulas are lightweight latency predictors; hardware calibration
compensates for implementation-dependent effects such as kernel fusion,
cache behavior, and routing overhead, rather than making these traffic
approximations exact.

\paragraph{Scope of the convex-cost assumption.}
The expert-touch term $U(s)$ introduces a saturation point at $s=E/K$.
Consequently, the Qwen MoE cost model need not be globally convex in
$s$, and the convex-cost premise of Proposition~\ref{prop:unimodal} must be
checked for the budget range in use. The dense-model argument does not by
itself establish a global first-decrease stopping guarantee for this MoE
extension; the MoE experiment reports the controller's empirical behavior.

\subsection{Calibration Cost}
\label{app:subsec:calib_cost}

\input{tables/appendix/calibration_cost}

\algo calibrates once per (GPU, target) pair. In the Qwen3-4B/A6000 setup of Table~\ref{tab:calibration_cost}, this takes $70.29$ seconds, dominated by profiling. Kernel-latency changes (e.g., GPU, target/draft model, block size, or attention backend) require recalibration; tasks, datasets, context lengths, and concurrency levels reuse the same calibration.

%% file: tables/appendix/calibration_cost.tex
\begin{table}[!htbp]
\centering
\captionsetup{position=top,skip=7pt}
\caption{\textbf{One-time calibration cost.} Qwen3-4B on a single NVIDIA A6000, measured over a $42$-point grid ($126$ timed cycles). The \emph{profiling sweep} times the verifier at each grid point; \emph{curve fitting} solves for the linear calibration coefficients.}
\label{tab:calibration_cost}
\small
\setlength{\tabcolsep}{6pt}
\renewcommand{\arraystretch}{1.05}
\begin{tabular}{@{}lr@{}}
\toprule
\textbf{Calibration stage} & \textbf{Cost} \\
\midrule
Profiling sweep & $70.27$\,s \\
Curve fitting & $0.02$\,s \\
\midrule
\textbf{Total} & $\mathbf{70.29}$\,\textbf{s} \\
\bottomrule
\end{tabular}
\end{table}

%% file: tex/appendix/06_experimental_setup.tex
\section{Detailed Experimental Setup}
\label{app:sec:detail_exp_setup}

\paragraph{Models and datasets.}
Our main evaluation uses three dense autoregressive target models:
Qwen3-4B, Qwen3-8B, and Llama-3.1-8B-Instruct~\citep{yang2025qwen3,grattafiori2024llama}.
Each target model is paired with its compatible block-diffusion drafter from
DFlash~\citep{chen2026dflash}. Our evaluation covers four domains. For
mathematical reasoning, we use GSM8K~\citep{cobbe2021trainingverifierssolvemath},
MATH500~\citep{hendrycks2021measuring}, and AIME25~\citep{aime25}. For code
generation, we use HumanEval~\citep{chen2021evaluating},
MBPP~\citep{austin2021program}, and LiveCodeBench~\citep{jain2024livecodebench}.
For general instruction following, we use Alpaca~\citep{alpaca} and
MT-Bench~\citep{zheng2023judging}. For long-context understanding, we use seven
English subsets of LongBench~\citep{bai2024longbenchbilingualmultitaskbenchmark}:
Qasper~\citep{dasigi-etal-2021-dataset}, MultiFieldQA-en, GovReport~\citep{huang-etal-2021-efficient},
MultiNews~\citep{fabbri-etal-2019-multi}, TriviaQA~\citep{joshi-etal-2017-triviaqa},
SAMSum~\citep{gliwa-etal-2019-samsum}, and PassageRetrieval-en.
Beyond these dense models, we also evaluate \algo on the mixture-of-experts
(MoE) target Qwen3-Coder-30B-A3B-Instruct, paired with its compatible DFlash
drafter, on HumanEval, MBPP, and LiveCodeBench. Appendix~\ref{app:subsec:moe}
provides the MoE experimental setup and results. Unless stated
otherwise, all experiments are run with batch size~1.
\looseness=-1

\vspace{-6pt}
\paragraph{Metrics and hardware.}
Our primary metric is speedup over target-only autoregressive decoding, computed as the ratio between target-only per-token latency and speculative-decoding per-token latency. We also report average accepted length AAL ($\tau$). We evaluate hardware adaptability on four NVIDIA GPUs: A100 (80\,GB HBM2e), H100 (80\,GB HBM3), RTX PRO 6000 Blackwell (96\,GB GDDR7), and RTX A6000 (48\,GB GDDR6). All runs use a single GPU, with batch size~1 except for the SGLang dynamic-batching study in Section~\ref{subsec:sglang}; the per-step controller overhead of \algo is negligible relative to the target forward pass on every tested machine.
\looseness=-1

\vspace{-6pt}
\paragraph{Baselines.}
For dense targets, we compare against DFlash, the most directly related block-diffusion speculative decoding baseline, and EAGLE-3~\citep{li2025eagle}, a strong tree-based speculative decoding baseline with an AR-style drafter. For EAGLE-3, we use the open-source AngelSlim~\citep{angelslim2026} implementation and set the tree size to $60$ by default. Appendix~\ref{app:subsec:eagle3_fairness} examines comparison fairness through EAGLE-3 hyperparameter sweeps and comparisons with \algo-Fixed at matched tree budgets. All baselines use the same target model, benchmark, decoding temperature, and batch size as \algo.
\looseness=-1

\vspace{-6pt}
\paragraph{Compute resources.}
A single (target, benchmark, method, GPU, temperature) configuration completes in roughly 4 hours on the eight short-context benchmarks and 5 hours on the LongBench English subsets, depending on output length and accepted-tokens-per-step. The dense-model experiment grid---three targets, four GPUs (with not all (model, GPU) cells run; see Tables~\ref{tab:main_res} and~\ref{tab:add_res_h100} and Figures~\ref{fig:additional_results},~\ref{fig:fixed_budget_compare}), three methods, eight short-context plus seven long-context benchmarks, two temperatures, and the budget sweep ($N\in\{32,64,128,256,512,1024\}$ plus \algo-Adaptive)---requires approximately 12 GPU-hours to reproduce the reported numbers. Development and ablation runs not included in the paper (drafter compatibility checks, latency-model calibration on configurations not reported, alternative scoring functions, and rerun cycles after pipeline fixes) contributed an additional ${\sim}$1.5--2$\times$ of that compute. No model training was performed: \algo is training-free, and all target and drafter checkpoints are used as released.
\looseness=-1